\documentclass{article}

\PassOptionsToPackage{square,numbers}{natbib} % or round, authoryear, etc.
\usepackage{natbib}
\usepackage{graphicx}
\usepackage{wrapfig}
\usepackage{lipsum}
\usepackage{multirow}
\usepackage{array,float}
\usepackage{amsmath} 
\usepackage{bm}
\usepackage{caption}
\usepackage{subcaption}
\usepackage[skins,breakable]{tcolorbox}

\floatstyle{plain} % or boxed, ruled, etc.
\newfloat{example}{htbp}{loe}
\floatname{example}{Example}

\usepackage[preprint]{neurips_2025}

\usepackage[utf8]{inputenc} % allow utf-8 input
\usepackage[T1]{fontenc}    % use 8-bit T1 fonts
\usepackage{hyperref}       % hyperlinks
\usepackage{url}            % simple URL typesetting
\usepackage{booktabs}       % professional-quality tables
\usepackage{amsfonts}       % blackboard math symbols
\usepackage{nicefrac}       % compact symbols for 1/2, etc.
\usepackage{microtype}      % microtypography
\usepackage{xcolor}         % colors
\usepackage{tcolorbox}

\title{Overclocking LLM Reasoning: Monitoring and Controlling  Thinking Path Lengths in LLMs}

\author{%
  Roy Eisenstadt \\
  \small Tel-Aviv University\\
  \small \texttt{royeisen@gmail.com} \\
  \And
  Itamar Zimerman \\
  \small Tel-Aviv University, IBM Research \\
  \small \texttt{zimerman1@mail.tau.ac.il} \\
  \And
  Lior Wolf \\
  \small Tel-Aviv University \\
  \small \texttt{wolf@cs.tau.ac.il}
}

\begin{document}

\maketitle

\begingroup
\renewcommand{\thefootnote}{} % remove number
\footnotetext{\small Project page: \url{https://royeisen.github.io/OverclockingLLMReasoning-paper}}
\addtocounter{footnote}{-1} % prevent counter from increasing
\endgroup

{
\vspace{-26px}
\begin{figure}[h!]
    \centering
\includegraphics[width=1.0\linewidth]{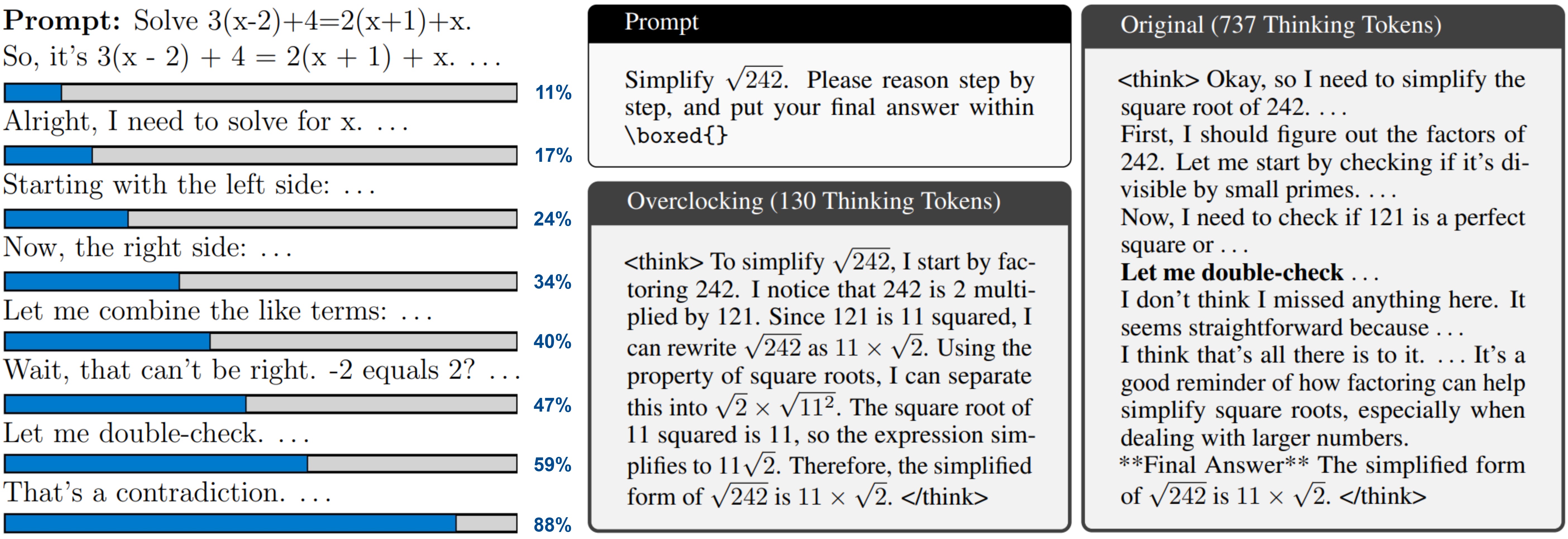}
\vspace{-0pt} % Optional: reduce space between figure and labels
\begin{minipage}{\linewidth}
    \centering
    \hspace*{-0.155\linewidth}\makebox[0pt][l]{{(a)}}
    \hspace*{0.495\linewidth}\makebox[0pt][l]{{(b)}}
\end{minipage}
\vspace{-9pt} % Optional: further tighten spacing
\caption{Applications of our method. (a) Monitoring the reasoning progress. (b) Overclocking it.}
\label{fig:teaser}
\vspace{-10pt}
\end{figure}%
\vskip 0.1in

}
% \begin{figure}[t]
% \centering
% \includegraphics[width=1.0\linewidth]{figures/mainFig.jpg}
% \vspace{-0pt} % Optional: reduce space between figure and labels
% \begin{minipage}{\linewidth}
%     \centering
%     \hspace*{-0.22\linewidth}\makebox[0pt][l]{{(a)}}
%     \hspace*{0.52\linewidth}\makebox[0pt][l]{{(b)}}
% \end{minipage}
% \vspace{-8pt} % Optional: further tighten spacing
% \caption{Two applications of our method. (a) Monitoring the progress of the reasoning process. (b) Overclocking the process.}
% \label{fig:teaser}
% \vspace{-14pt}
% \end{figure}

\begin{abstract}
Recently, techniques such as explicit structured reasoning have demonstrated strong test-time scaling behavior by enforcing a separation between the model’s internal ``thinking" process and the final response. A key factor influencing answer quality in this setting is the length of the thinking stage. When the reasoning is too short, the model may fail to capture the complexity of the task. Conversely, when it is too long, the model may overthink, leading to unnecessary computation and degraded performance. This paper explores and exploits the underlying mechanisms by which LLMs understand and regulate the length of their reasoning during explicit thought processes. First, we show that LLMs encode their progress through the reasoning process and introduce an interactive progress bar visualization, which is then used to reveal insights on the model's planning dynamics. Second, we manipulate the internal progress encoding during inference to reduce unnecessary steps and generate a more concise and decisive chain of thoughts. Our empirical results demonstrate that this ``overclocking'' method mitigates overthinking, improves answer accuracy, and reduces inference latency. %In particular, we show that it improves the performance of DeepSeek-R1-Qwen-32B by 60\% on the Math-500 dataset, while reducing the number of reasoning steps by approximately 30\%. 
Our code is publicly available.

{\small
\vspace{0.5em}
\hspace{.5em}
\includegraphics[width=1.25em,height=1.15em]{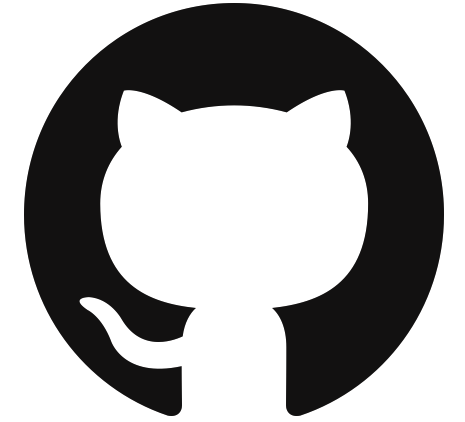}\hspace{.75em}
\parbox{\dimexpr\linewidth-7\fboxsep-5\fboxrule}{\url{https://github.com/royeisen/reasoning_loading_bar}}
\vspace{-.9em}
}

\end{abstract}

\vspace{-11pt}
\section{Introduction}
\vspace{-3pt}

In recent years, Large Language Models (LLMs) such as ChatGPT \citep{openai2023chatgpt} have demonstrated remarkable capabilities across both general-purpose and domain-specific challenges, exhibiting surprising generalization abilities~\citep{kojima2022large}. An emerging frontier for enhancing their performance is test-time scaling, which aims to improve model responses by dynamically allocating additional computational effort during inference~\citep{snell2024scaling,welleck2024decoding}. %This process takes inspiration from human behavior, where individuals often spend time thinking before responding to non-trivial questions, and is designed to strengthen the model’s ability to perform complex reasoning steps.

A prominent strategy for enabling effective long-form reasoning is explicit structured reasoning, which separates the model’s reasoning process from its final answer~\citep{wei2022chain}. This is commonly implemented using special tokens that mark the start and end of the reasoning phase (e.g., <think> and </think>, which are employed by DeepSeek-R1~\citep{guo2025deepseek}), encouraging the model to deliberate before responding.%producing a response.

The effectiveness of such techniques depends heavily on the length of the thinking stage~\citep{jin-etal-2024-impact,wu2025more}. Too few steps may fail to capture sufficient task complexity, while too many can lead to overthinking and unnecessary computation~\citep{sui2503stop}. Thus, controlling the thinking path length and understanding its underlying mechanisms are essential for balancing accuracy, efficiency, and responsiveness.

While several prior works have explored the impact of reasoning length on model quality and proposed techniques to control or optimize response length through specialized training procedures~\citep{aggarwal2025l1} or prompting~\citep{wu2025effectively}, they do not investigate the underlying mechanistic design that governs how reasoning length is determined. Moreover, none of these approaches address the unique case of structured reasoning, where the model explicitly separates a holistic thinking stage from the final response. We consider this setting particularly important, as it offers a valuable opportunity to better understand and exploit the ``thinking phase" of reasoning models.

This work tackles both of these gaps. First, we investigate whether LLMs are capable of monitoring their relative position within the % overall
thinking process. Possessing even an implicit form of progress monitoring may enable aspects of key cognitive concepts such as self-regulation~\citep{zimmerman2002becoming} and metacognition~\citep{flavell1979metacognition}. To identify mechanisms that encode this information, we perform a regression analysis, and show that the relative position can be captured by projections that we term ``progress vectors".

The extracted information is then used to create an interactive loading bar visualization, see Figure~\ref{fig:teaser}(a) that depicts the model’s progress throughout the thinking phase, making the reasoning process more transparent and easier for users to collaborate with.

The ability to extract progress information does not mean that the model employs it mechanistically, unless an intervention analysis is performed. We thus manipulate the internal representation along the progress vectors and achieve a clear modulation of the length of the thinking phase, showing overclocking effects. The former is depicted in Figure~\ref{fig:teaser}(b). Reassuringly, this modulation does not tend to be detrimental to the LLM's performance. In fact, we show that overclocking can improve the model's performance by mitigating overthinking, enhancing computational efficiency, and tailoring the model’s reasoning depth to each task’s complexity.

% \begin{figure}
%     \centering
%     \includegraphics[width=1.0\linewidth]{figures/LLMloadingbar.jpg}
%     \caption{\lw{SEPARATE TO PANELS 
%     (a) (b) THE FIB EXAMPLE IS NOT GOOD SINCE JUST GIVING THE RESULT IS NOT GOOD. ALSO THE VISUALIZATION ON THE LEFF IS TERRIBLE} Placeholder. Loading bar and qualitative Examples of lengh-control}
%     \label{fig:enter-label}
%     \vspace{-18pt}
% \end{figure}

\textbf{Our main contributions are as follows:} (i) We provide the first empirical evidence that LLMs maintain an internal estimate of their relative position within the explicit thinking phase. This finding sheds light on the plausibility of planning and self-monitoring abilities of LLMs, concepts typically associated with cognitive capabilities. (ii) We identify an internal encoding of this information by learning  progress vector projections, and employ these projections to expose a dynamic thinking progress bar. (iii) We perform an intervention study and manipulate the progress vectors to overclock and downclock the reasoning process. Finally, 
(iv) We empirically demonstrate that interventions of the progress vectors improve both the efficiency and the  effectiveness of strong LLMs such as DeepSeek-R1 by  mitigating overthinking and reducing the generation of unnecessary reasoning steps.

\section{Related Work}

%Here we describe the scientific context and introduce the key terminology for discussing our method.
% This section describes the scientific context and introduces the necessary terminology used to discuss our method.

{\bf Test-time scaling} refers to a recent trend in which the reasoning depth of an LLM is dynamically increased during inference, allowing the model to perform complex multi-step reasoning and extend its problem-solving capabilities~\citep{snell2024scaling,welleck2024decoding}. This approach heavily relies on the model’s ability to reason effectively over long trajectories. One notable technique for improving the effectiveness of multi-step reasoning is Chain-of-Thought (CoT) prompting~\citep{wei2022chain}, which encourages the model to generate a sequence of intermediate reasoning steps before producing its final answer by incorporating additional guidance into the prompt. This technique has been adopted and extended in models such as DeepSeek-R1~\citep{guo2025deepseek}, OpenAI's O1~\citep{jaech2024openai}, and S1~\citep{muennighoff2025s1} which are fine-tuned on datasets containing CoT-style reasoning steps prior to the final answer. A further refinement was proposed by DeepSeek-R1~\citep{guo2025deepseek}, which explicitly enforces open-ended structured reasoning through the use of dedicated <think> and </think> tokens that mark the beginning and end of the model's ``thinking'' phase. 

{\bf The length of the reasoning chain} is a key factor influencing the performance of modern reasoning models. If the chain is too long, the model may engage in unnecessary computation and suffer from overthinking~\citep{sui2503stop,su2025between,chen2024not}, a phenomenon in which excessive reasoning steps degrade answer quality. On the other hand, if the reasoning is too short, the model may fail to capture the complexity required to solve the task effectively~\cite{wang2025thoughts}. To address this trade-off, state-of-the-art LLMs are increasingly designed to support adaptive thinking, where the length of the reasoning phase is dynamically adjusted based on task complexity. Two main strategies for improving the length-control the duration of a model's reasoning have been proposed: (i) Inference-time methods modify the prompt or inject guidance while the model is generating its chain of thought. Notable examples include~\citet{kojima2022large}, who optimize the initial prompt, and~\citet{wu2025effectively,jin2024impact}, who dynamically intervene during reasoning by prompting to shorten or lengthen the reasoning trajectory (ii) Training-time methods fine-tune the model with objectives and datasets that explicitly encourage regulation of the reasoning length. A notable example is Length-Controlled Policy Optimization (LCPO)~\citep{aggarwal2025l1}, which frames reasoning-length control as a RL objective. Beyond the context of controlling reasoning-chain length, length control has long been studied in standard LLMs to manage overall output length. A substantial body of work addresses this objective across applications such as summarization~\citep{kwon-etal-2023-abstractive,retkowski2024zero}. In contrast to these methods that primarily aim to control the length of the generated answer for efficiency or user-specific preferences, our work focuses on the thinking length rather than the final output. This allows us to gain insights into the model’s planning, monitoring, and control capabilities, which we then leverage to improve model quality and model responsiveness.

{\bf Mechanistic interpretability} aims to reverse-engineer LLMs so they can be understood, trusted, and controlled~\citep{elhage2021mathematical,olah2020zoom}. %Although modern LLMs are widely applied, they are still considered black boxes, and the lack of mechanistic insight further limits broader adoption. 
This is done by connecting observable behaviors to concrete neural components by assigning semantic roles to individual neurons or weights and isolating compact ``circuits''% that implement specific functions
. Prominent examples include the induction-heads%circuit
~\citep{olsson2022context} and mechanisms %responsible
for arithmetic~\citep{kantamneni2025language,nanda2023progress,zhong2023the,chughtai2023toy} and in-context learning via task-vectors~\citep{hendel2023context}.%, and positional awareness without PE~\cite{chi2023latent}.
Our work follows this line of research and aims to uncover the intrinsic mechanisms that encode a model’s relative position within its internal reasoning process, with the goal of better understanding and even controlling the planning abilities of reasoning LLMs.

\section{Method\label{sec:method}}

We investigate monitoring and control of the thinking phase in reasoning models. In Section~\ref{subsec:MethodMonitoring}, we show that the model’s hidden representations encode information about its position within the thinking phase. This allows for real-time monitoring in the form of a loading bar and reveals distinct thinking patterns. In Section~\ref{subsec:MethodControl}, we demonstrate that this internal mechanism can be modified to control the depth of the thinking phase.

\subsection{Monitoring the Thinking-Phase \label{subsec:MethodMonitoring}}

We hypothesize that as part of learning to reason effectively, models must also implicitly learn to track their progress through the thinking phase, maintaining an estimate, for example, of how close they are to the final answer. 
Since progress-tracking is input-dependent, such information cannot be stored in the model’s static weights and must instead be encoded dynamically in the hidden representations passed between layers. We chose to extract information from the final hidden layer for two main reasons. First, this layer provides the richest token-level representations, having integrated information from all previous layers. Second, our goal is to eventually intervene in the reasoning process by manipulating the thinking progress. The final hidden layer is especially suitable for this because it is close to the token embedding space, allowing for effective intervention. Moreover, it lies downstream of the attention layers, which cache intermediate states. This means that modifying the hidden representation at this stage can influence the immediate output token without disrupting the model’s internal memory, thereby isolating the effect of the intervention to a single step.

As another design choice, we focus on models that perform explicit structured reasoning, characterized by a {distinct and continuous} reasoning phase delimited by \texttt{<think>} and \texttt{</think>} tokens, as employed in DeepSeek-R1~\citep{guo2025deepseek}. This structure enables us to quantify the model's progression within the thinking phase by precisely labeling each token with an interpolated value between zero and one, according to its relative position. 

To learn a thinking progress predictor, we initially sample entries from a dataset of mathematical problems. We then prompt a reasoning model with a single problem each time, generating an answer that contains a thinking phase. For a generation that ended successfully, we specifically observe its thinking trajectory \( T = w_{1}w_{2} \ldots w_{N} \), which is the sequence of tokens between ``\texttt{<think>}'' and ``\texttt{</think>}'', and collect the representations from the last layer, \( \{ \mathbf{h}_{j}^{(k)} \in \mathbb{R}^{d} \mid j = 1, \ldots, N_k \} \) for each token $j$ in the $k$ trajectory, where \(d\) is the hidden size of the reasoning model and \(N_k\) is the number of tokens in that trajectory. We pair each hidden state with its normalized position in the sequence as a regression label.

Formally, we construct a dataset $D$ by:
\begin{equation}
\mathcal{D} = \{ ( \mathbf{h}_{j}^{(k)},\ p_{j}^{(k)} ) \;|\; k = 1, \ldots, K,\ j = 1, \ldots, N_k \} \,,
\end{equation}
where \( \mathbf{h}_{j}^{(k)} \in \mathbb{R}^d \) is the hidden representation of the \(j\)-th token in the \(k\)-th thinking trajectory, and \( p_{j}^{(k)} = \frac{j}{N_k} \in (0, 1] \) is the token's relative position in its sequence. Here, \(K\) denotes the number of sampled trajectories and the total number of samples in \(\mathcal{D}\) is \(M = \sum_{k=1}^K N_k\).

This allows us to optimize for a progress extraction function \(f_{\theta}\) which maps the hidden representations to their relative positions in the form of a regression task 
%
%
%
%\begin{equation}
$\theta^{*} = \arg\min_{\theta} \sum_{(\mathbf{h}, p) \in \mathcal{D}} ( f_{\theta}(\mathbf{h}) - p )^2$.
%\end{equation}

%\begin{wrapfigure}{rb}{0.48\linewidth}
% \vspace{-14pt}
% \begin{figure}[t]
%     \centering
%     \includegraphics[width=0.699\linewidth]{figures/linear_vs_mlp_regressor.png}
%     \caption{\small Linear vs Non-linear regression MSE loss for the task of monitoring thinking phase. For each model-dataset pair 30 problems were sampled.}
%     \label{fig:lr_vs_ffn}
% \end{figure}

\begin{figure}[t]
    \centering
    \begin{minipage}[t]{0.49\linewidth}
        \centering
        \includegraphics[width=\linewidth]{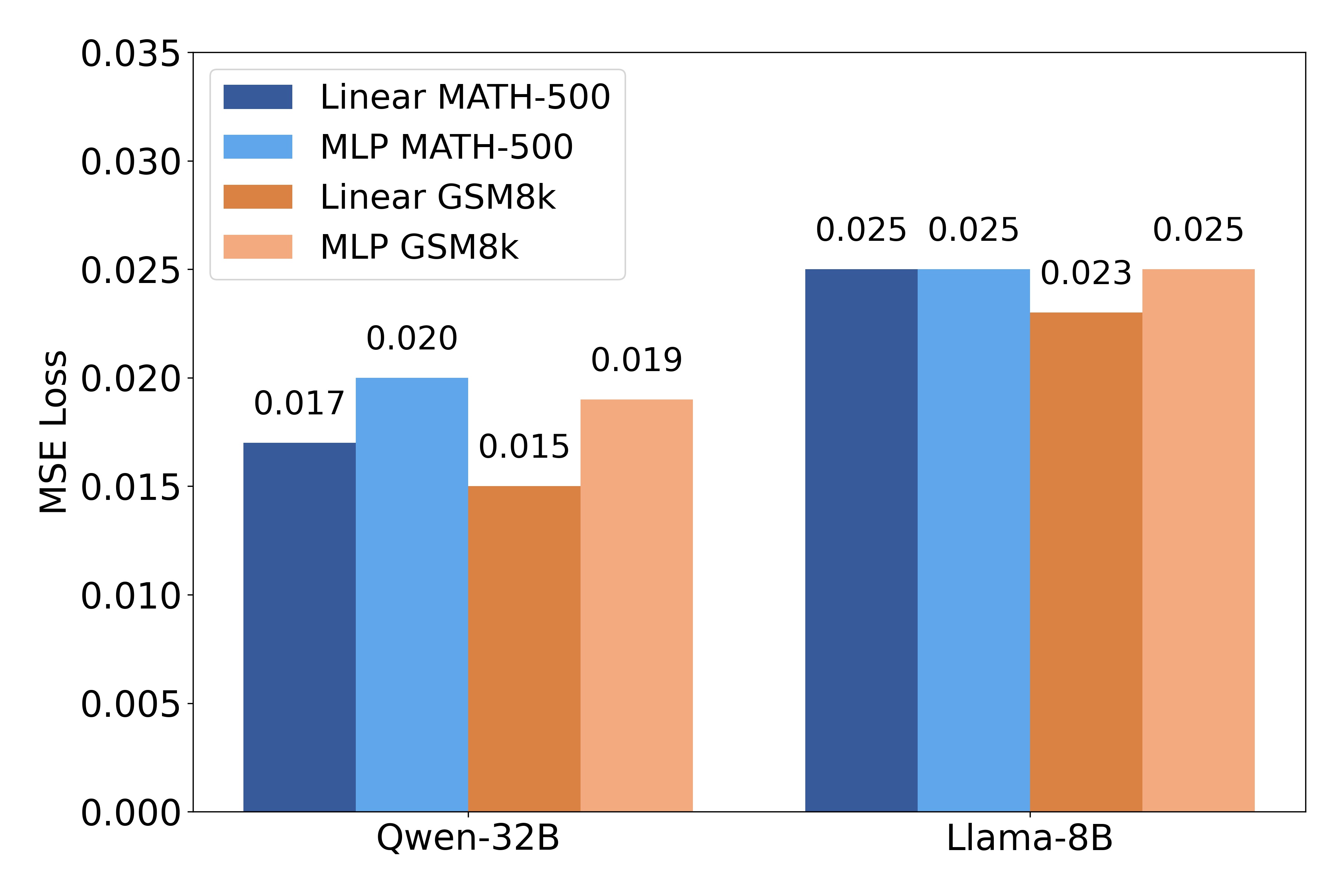}%linear_vs_mlp_regressor.png}
        \caption{Linear vs Non-linear regression MSE loss for monitoring the thinking phase. For each model-dataset pair, 30 problems were sampled.}
        \label{fig:lr_vs_ffn}
    \end{minipage}\hfill
    \begin{minipage}[t]{0.49\linewidth}
        \centering
        \includegraphics[width=\linewidth]{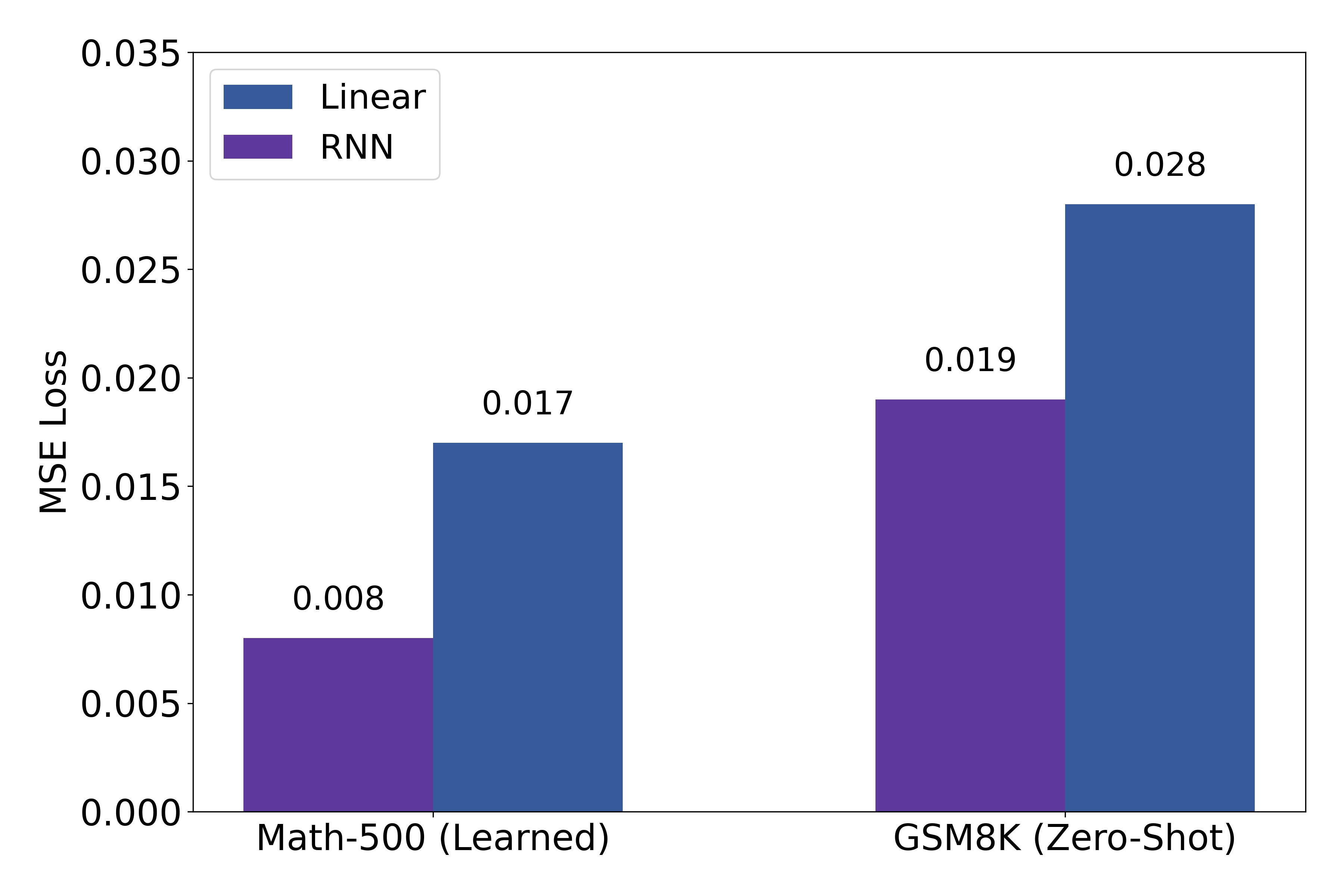}%{figures/zeroshotRNN.jpeg}
        \caption{{MSE of Linear vs RNN for thinking-phase monitoring regression in learned and zero-shot settings.}}
        \label{fig:lr_vs_rnn}
    \end{minipage}
\end{figure}
%    \vspace{-8pt}
%\end{wrapfigure}
%
We conduct experiments using two reasoning models - DeepSeek-R1-Distill-Qwen-32B and DeepSeek-R1-Distill-Llama-8B\cite{guo2025deepseek} - both of which are designed to explicitly separate the thinking process from final answers. We evaluate them on two benchmark datasets of mathematical problems: Math500\cite{lightman2023lets} and GSM8k~\cite{cobbe2021gsm8k}. For each dataset, we sample 30 problems and prompt each model to generate 5 distinct responses per problem using a temperature of 0.6. The prompt format includes the instruction: “Please reason step by step, and put your final answer within \texttt{\textbackslash boxed\{\}}. \texttt{<think>}”, as recommended by \citet{guo2025deepseek}. Following the dataset curation procedure described earlier, we obtain four distinct progress regression datasets, each corresponding to a unique combination of reasoning model and problem set. We split each dataset into train and test sets using an approximate 80/20 ratio based on entire thinking trajectories, ensuring that all generations for a given problem remain within the same split.

We first fit a linear regressor parametrized by \(\bm{\theta} \in \mathbb{R}^{d}\) as the function $f_\theta$ for estimating the progress property $\bar{p}^{(i)}_{j}  = \bm{\theta}^T \mathbf{h}^{(i)}_j$. We 
%
%\begin{equation}
%\bar{p}^{(i)}_{j}  = \bm{\theta}^T \mathbf{h}^{(i)}_j\,,
%\end{equation}
 refer to the parameter vector $\bm{\theta}$ as the \textit{``thinking progress vector''} (\textbf{TPV}).

To assess whether the estimated progress $\bar{p}^{(i)}_{j}$ can be more effectively captured by a more complex model, we compare a simple two-layer feedforward network (FFN) with our proposed TPV model. As shown in Figure~\ref{fig:lr_vs_ffn}, both models achieve low loss on the test sets, with no observable improvement from the FFN. Guided by the principle of favoring simplicity, and supported by these empirical findings, we select TPV as our method of choice for progress extraction in subsequent experiments.

\begin{figure}[t]
    \centering
    \setlength{\tabcolsep}{4pt}
    \renewcommand{\arraystretch}{1.0}
    \begin{tabular}{@{}m{0.5\linewidth}@{} m{0.5\linewidth}@{}}
        % Left column: (a)
        \hspace{10pt}
        \begin{tabular}{@{}c@{}}
          \includegraphics[width=1.0\linewidth, trim=0pt 18pt 0pt 0pt, clip]{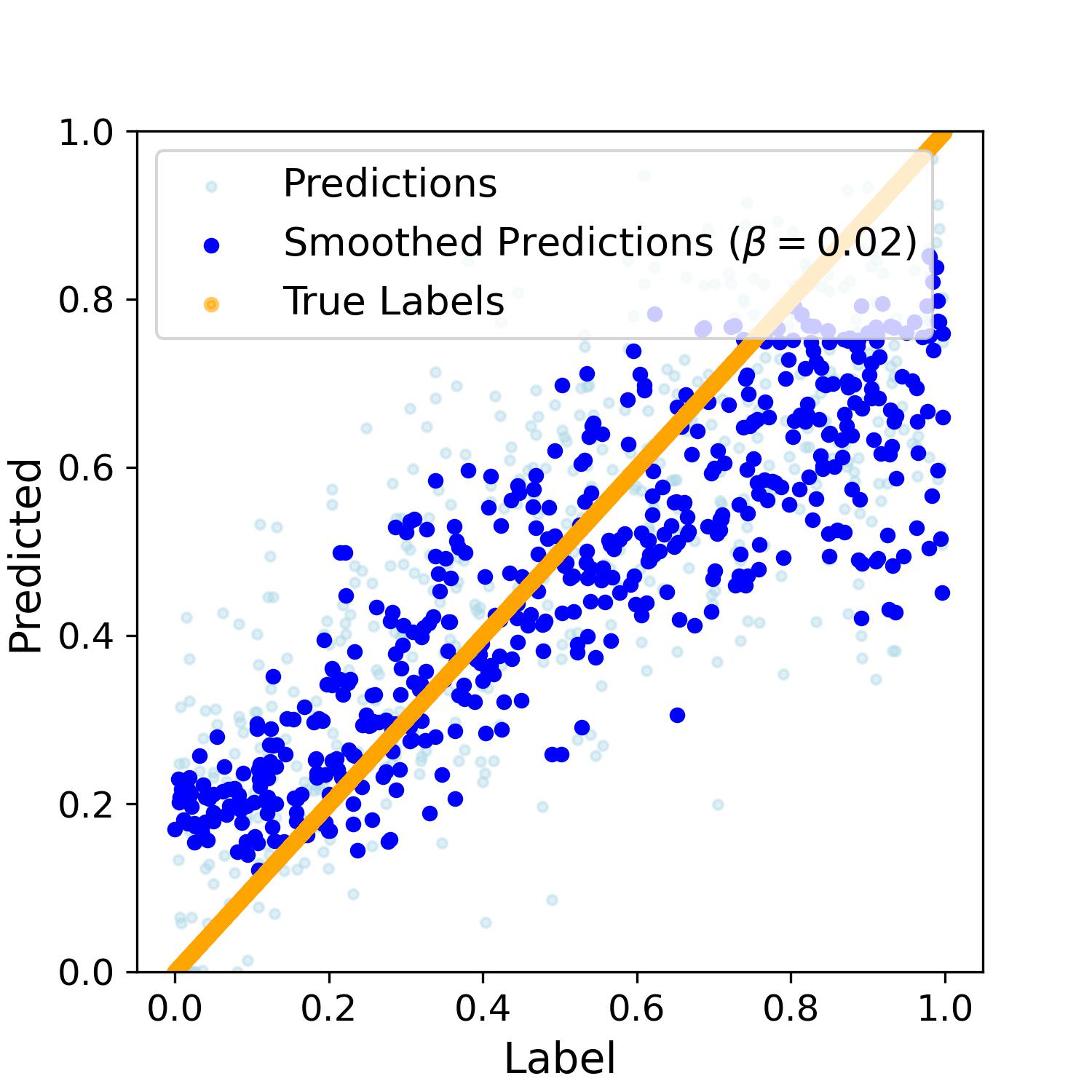}\vspace{-4pt}\\
            \small (a)
        \end{tabular}
        & \vspace{18pt}
        % Right column: (b) and (c)
        \hspace{18pt}
        \begin{tabular}{@{}c@{}}
            \includegraphics[width=0.89\linewidth, trim=0pt 15pt 0pt 0pt, clip]{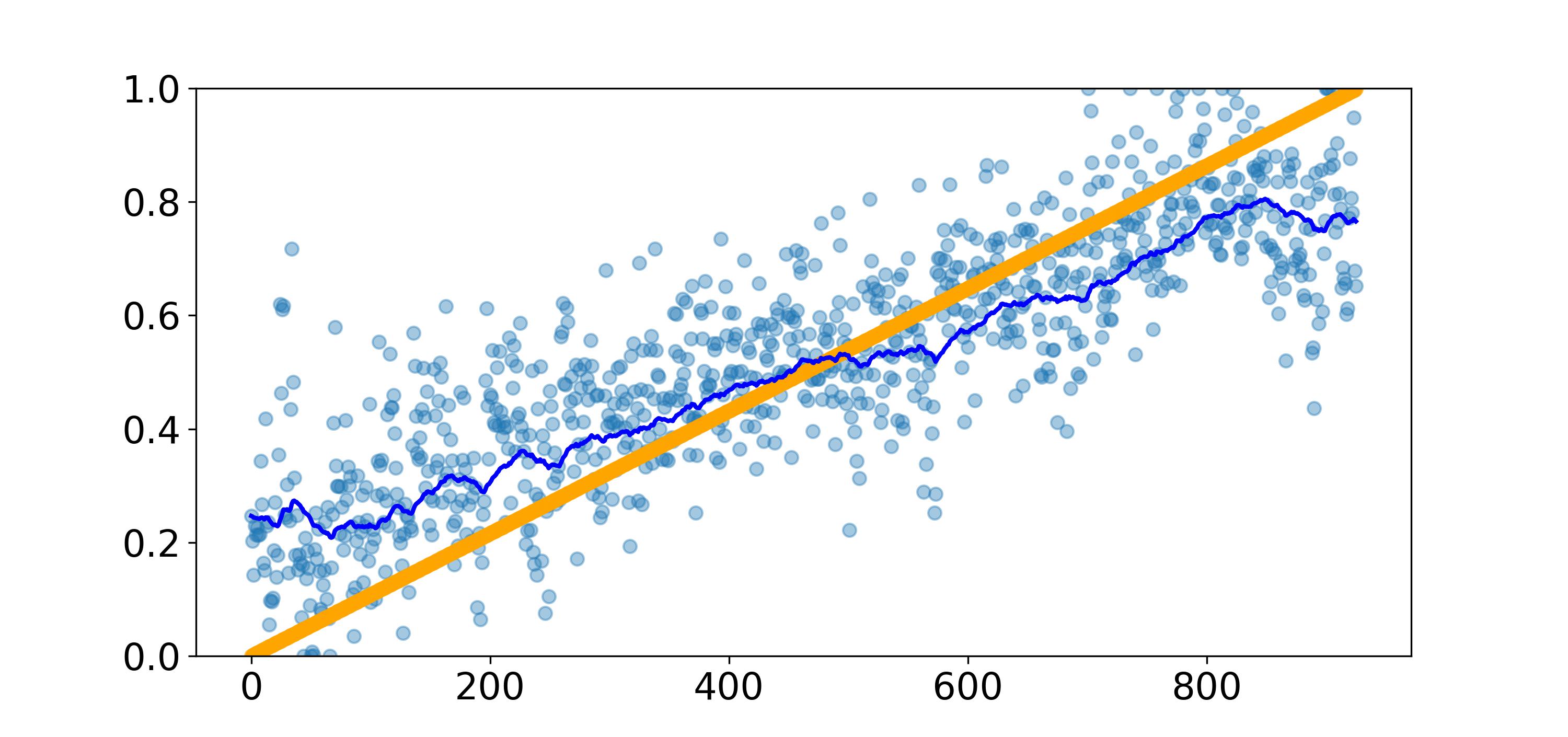} \\
             {\vspace{-2pt}}\small (b) \\
            {\vspace{-2pt}
            \includegraphics[width=0.89\linewidth, trim=0pt 15pt 0pt 0pt, clip]{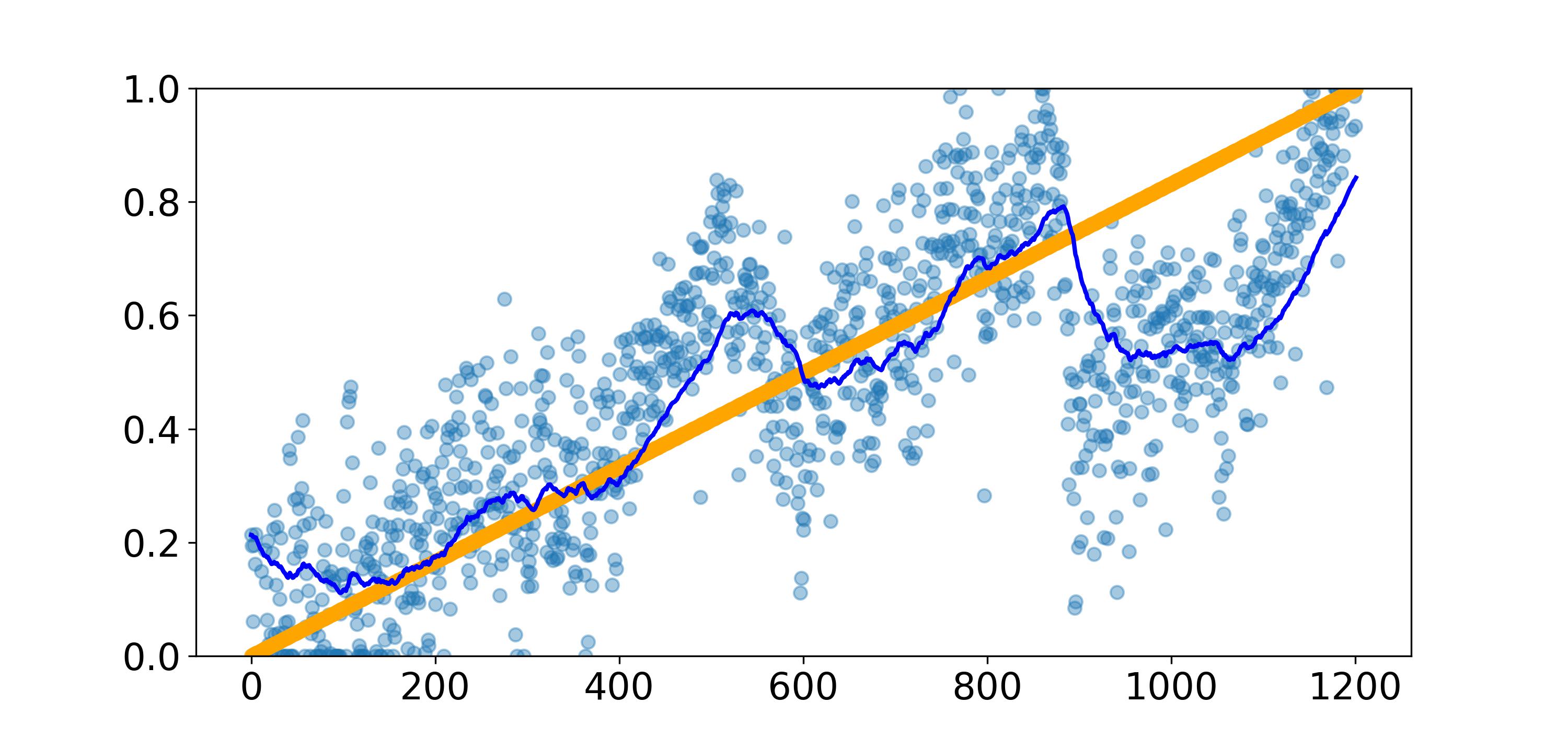}} \\
            \small (c)
        \end{tabular}
    \end{tabular}
    % \vspace{-6pt}
    \caption{\textbf{Relative Progress Prediction Analysis:}  
    (a) We sample 1000 random tokens from the thinking trajectories across all examples in the test set of the Math-500 dataset. The x-axis denotes the ground-truth relative position within the thinking phase, while the y-axis shows the predicted progress.  
    (b, c) Two representative examples from the Math500 test set, illustrating how the predictor estimates progress across the full thinking trajectory of a single generation. The x-axis denotes the decoding step index, and the y-axis shows the predicted value in the \([0,1] \) range.}
    \label{fig:regression_loading_Bar}
    % \vspace{-12pt}
\end{figure}

To improve the predictions, we leverage the model’s autoregressive nature, in which tokens are generated sequentially, and apply exponential smoothing over the prediction history to reduce noise. In Figure~\ref{fig:regression_loading_Bar}, we illustrate TPV predictions over problems from the Math-500 test set \cite{lightman2023lets}. The light blue and dark blue points represent the predictions and smoothed predictions correspondingly, while the orange line depicts the ideal loading bar obtained through linear interpolation. Figure~\ref{fig:regression_loading_Bar}(a) presents an aggregative view over data points from several thinking trajectories, while  Figures~\ref{fig:regression_loading_Bar} (b, c) showcase TPV predictions and smoothed predictions along thinking trajectories of single problems in the Math-500 test set. As shown in the figures, both approaches, with and without smoothing, successfully predict the relative position, while the latter produces more precise results that can be used to create a clearer, more interpretable loading bar.

%\begin{wrapfigure}{r}{0.455\linewidth}
% \vspace{-23pt}
% \begin{figure}[t]
%     \centering
% \includegraphics[width=0.691\linewidth]{figures/zeroshotRNN.jpeg}
% \vspace{-8pt}
%     \caption{\small{MSE of Linear vs. RNN for thinking-phase monitoring in learned and zero-shot settings.}}
%     \label{fig:lr_vs_rnn}
%     \end{figure}
%    \vspace{-27pt}
%\end{wrapfigure}
%
%

\begin{figure}[t]
    \centering
    \setlength{\tabcolsep}{2pt}
    \renewcommand{\arraystretch}{1.0}
    \begin{tabular}{@{}cc@{}}
        \begin{tabular}{@{}c@{}} 
            \includegraphics[width=0.485\linewidth]{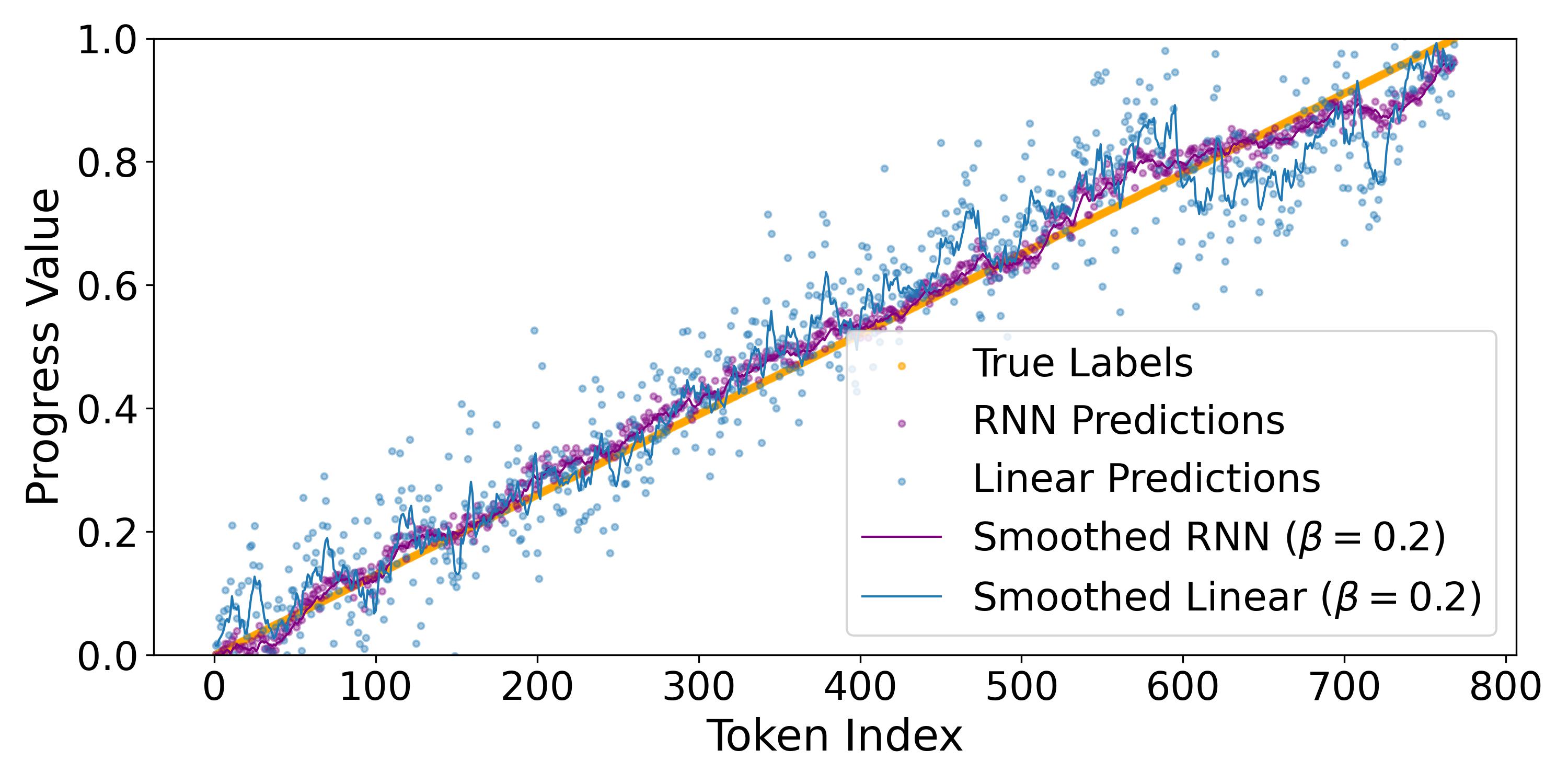} \\ \vspace{-2pt}
            \includegraphics[width=0.485\linewidth]{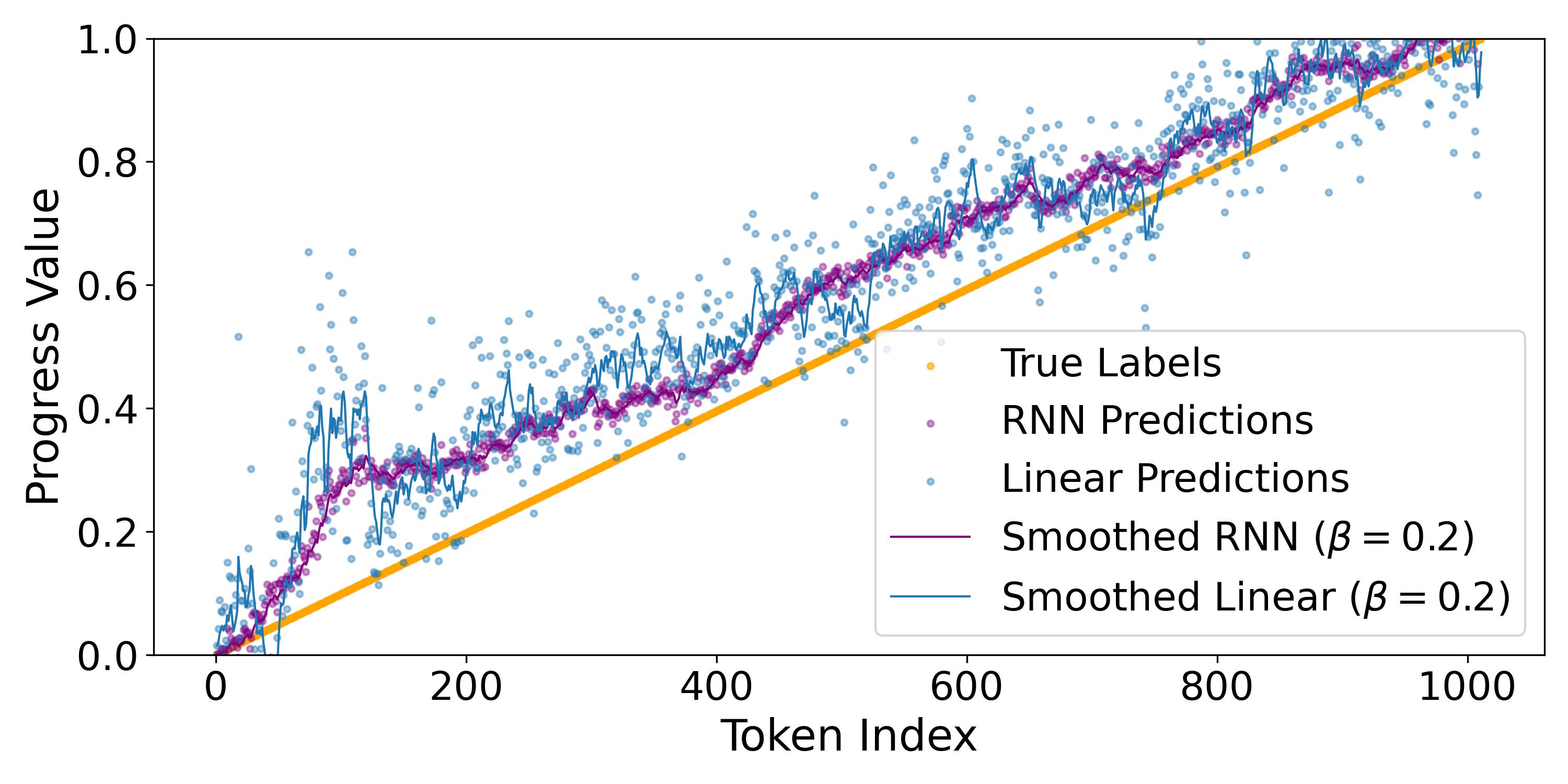} \\
            \small (a) MATH-500
        \end{tabular} &
        \begin{tabular}{@{}c@{}} 
            \includegraphics[width=0.485\linewidth]{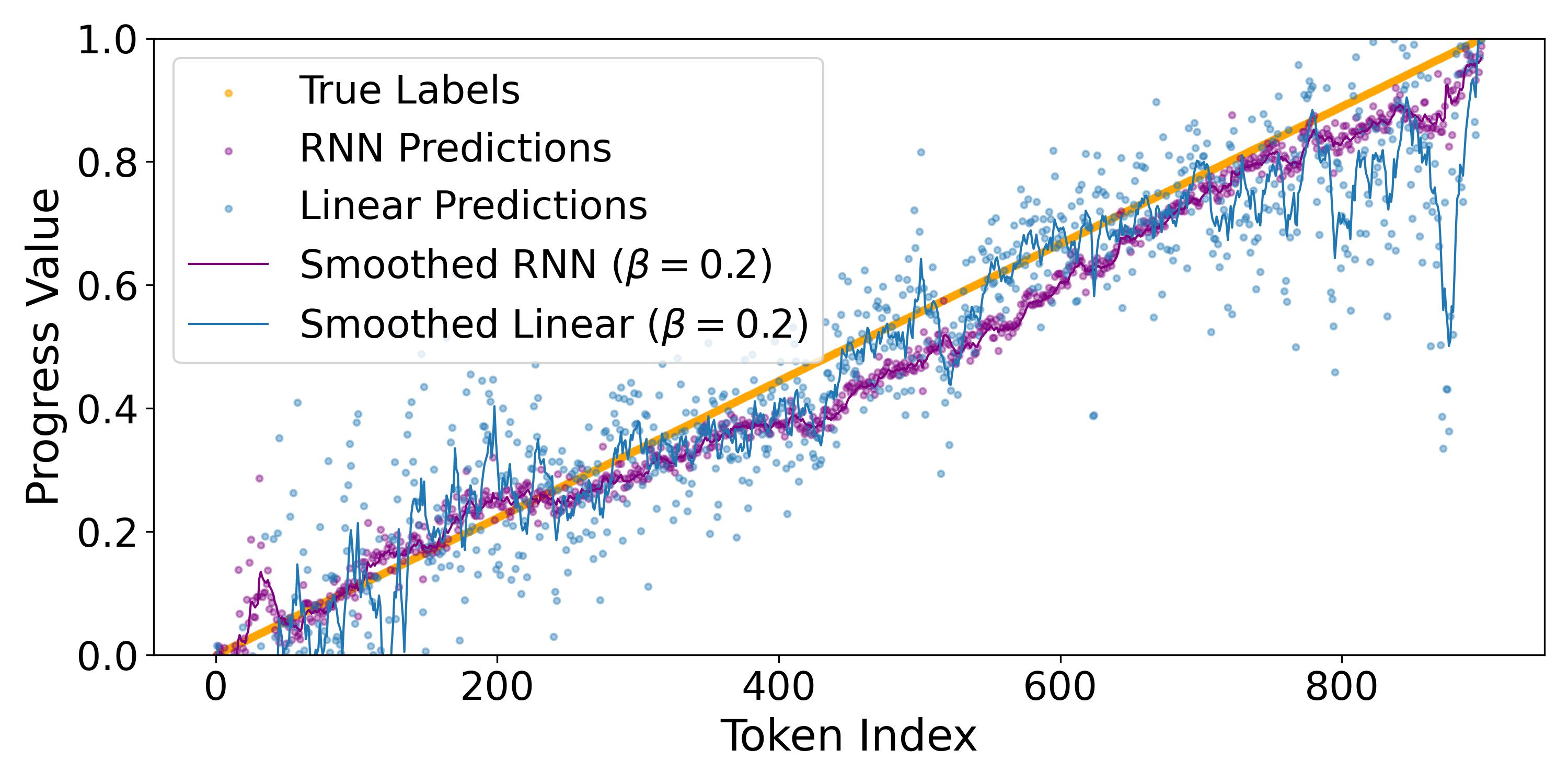} \\ \vspace{-2pt}
            \includegraphics[width=0.485\linewidth]{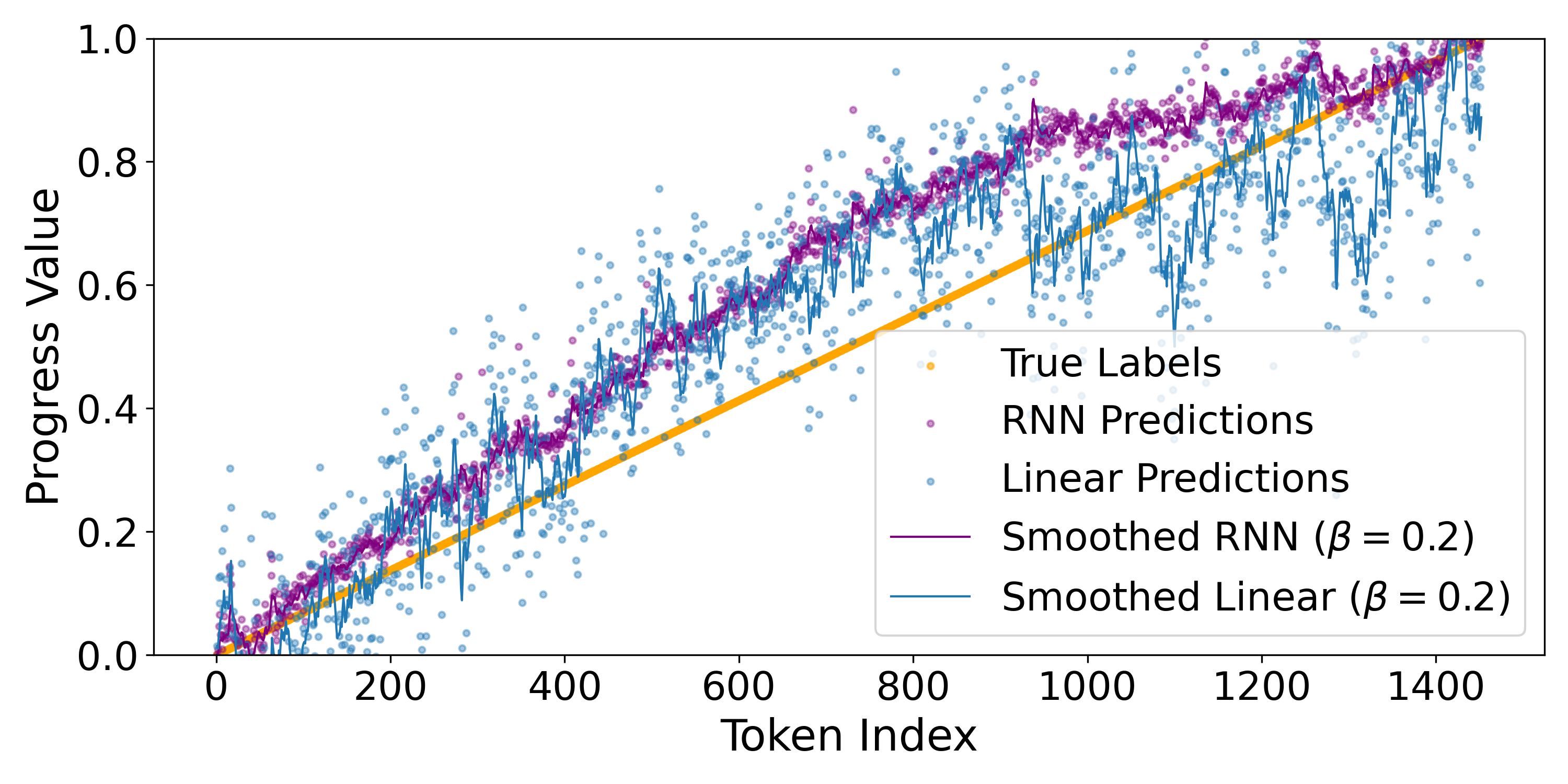} \\
            \small (b) GSM8K (zero-shot)
        \end{tabular} \\
    \end{tabular}
    \vspace{-2pt}
    \caption{{\textbf{Qualitative analysis of RNN-based progress prediction:} Predictions across time steps during the thinking phase.
    (a) Results on MATH-500 (in-domain); (b) results on GSM8K (zero-shot).}}
    \label{fig:loading_bar_rnn_zeroshot}
    % \vspace{-5pt}
\end{figure}

Motivated by this observation, and to better exploit the temporal structure of the progress-bar prediction task, we replace exponential smoothing with a trainable sequence model. Concretely, we adopt a single-layer GRU, parameterized by $\theta_\text{RNN}$, whose hidden and input dimensions are both set to $d$, and whose output at each step is a scalar. The network is trained on the sequence dataset $\mathcal{D'}$, using the same training samples as $\mathcal{D}$, only with a sequence of relative positions rather than performing single-step predictions: %
\begin{equation}
    \mathcal{D'} = \{ ( \mathbf{h}_{1}^{(k)}, \ldots, \mathbf{h}_{N_k}^{(k)} ) , (p_{1}^{(k)}, \ldots, p_{N_k}^{(k)} ) \mid k = 1, \ldots, K \ \}\,.
\end{equation}
% $$  \mathcal{D'} = \{ ( \mathbf{h}_{1}^{(k)}, \ldots, \mathbf{h}_{N_k}^{(k)} ) , (p_{1}^{(k)}, \ldots, p_{N_k}^{(k)} ) \mid k = 1, \ldots, K, \}
% $$
%$\mathcal{D'} = \{ ( \mathbf{h}_{1}^{(k)}, \ldots, \mathbf{h}_{N_k}^{(k)} ) , (p_{1}^{(k)}, \ldots, p_{N_k}^{(k)} ) \mid k = 1, \ldots, K \}$, 
% using the same training samples as $\mathcal{D}$, only with a sequence of relative positions rather than performing single-step predictions. 
To extend our analysis, we train the model on the MATH-500 training set and evaluate it on both the MATH-500 and GSM8K test sets, thereby covering both the learned and zero-shot regimes. The results are presented in Figure~\ref{fig:lr_vs_rnn}, showing that the RNN consistently achieves lower loss than the linear model in both settings. Although performance drops slightly in the zero-shot regime, the loss remains relatively low, suggesting that the model generalizes well across datasets. %
%
%While there is minor degradation in the zero-shot setting, the loss remains relatively low, suggesting that the model generalizes well across datasets. % enabling it to capture temporal dynamics and maintain positional awareness.
Finally, Figure~\ref{fig:loading_bar_rnn_zeroshot} presents a qualitative analysis of the RNN’s predictions over time during the thinking phase. %
%To further illustrate the impact of the RNN, Figure~\ref{fig:loading_bar_rnn_zeroshot} presents a qualitative analysis showing the RNN's predictions across time steps during the thinking phase.
In both learned and zero-shot regimes, the RNN predictions (shown in purple) produce outputs that are smooth, consistent over time steps, monotonically increasing, and accurately reflect the progress of the thinking phase, especially when combined with additional exponential smoothing.

% \vspace{-3pt}
% \paragraph{Semantic Meaning} Because the ``progress probe'' is obtained by projecting the final-layer representation into the output embedding space, one might expect its vector to resemble the embeddings of related tokens such as \texttt{<think>}, \texttt{<answer>}, or similar markers. To test this hypothesis, we empirically searched for the closest token embeddings in the output space and found no distinctive matches or relevant neighbors. This suggests that the TPV mechanism is not an artifact of the token-embedding geometry or of any specifications in DeepSeek's training data for enhanced reasoning.

% %\begin{figure}[h]
% \begin{wrapfigure}{r}{0.45\linewidth}
%     \vspace{-10pt}
%     \includegraphics[width=1.0\linewidth]{figures/loading_bar_regression.png}
%     \caption{\textbf{Intervention} PlaceHolder}
%     \vspace{-10pt}
% \label{fig:reliabilityProgressVectors}
% \end{wrapfigure}
% % \end{figure}

\subsection{Controlling Thinking Progress\label{subsec:MethodControl}}
A crucial question that arises is whether TPVs reflect a fundamental mechanism that the model uses to track its reasoning progress, or if they are merely residual artifacts that correlate with progress but do not play a causal role in the computation. To address this, we conduct intervention experiments.

\paragraph{Intervention Technique.}
%To conduct intervention experiments, we first define the intervention methodology. Building on our identification of the underlying mechanism via TPV, we recall the regression weight vector $\mathbf{w} \in \mathbb{R}^d$ and let $\mathbf{h} \in \mathbb{R}^d$ denote the representation from the final transformer block. The estimated relative position $\bar{p}$ is then given by $\bar{p} = \mathbf{w}^T \mathbf{h}$. To intervene in this mechanism, we add a direction vector $\mathbf{u} \in \mathbb{R}^d$ scaled by a coefficient $\alpha \in \mathbb{R}$, yielding a modified ``progress value'' $\bar{p_{\delta}}$. It is easy to show that the value of $\mathbf{u}$ that maximizes $\bar{p_{\delta}}$ is $\mathbf{w}$. We end up with the following technique:

The intervention experiments shift the hidden representation $\mathbf{h}$ in the direction of projection vector $\bm{\theta}$ by an amount $\alpha$: $\mathbf{h}_{\alpha} = \mathbf{h} + \alpha \bm{\theta}$. 
% %
% \begin{equation}
% \mathbf{h}_{\alpha} = \mathbf{h} + \alpha \bm{\theta} 
% \end{equation}
% %
The altered representation has a new prediction value:  $\bm{\theta}^{T}\mathbf{h}_{\alpha} = \bm{\theta}^{T} (\mathbf{h} + \alpha \bm{\theta} ) = \bar{p} + \alpha||\bm{\theta}||^{2} $.
%
% \begin{equation}
%     \bm{\theta}^{T}\mathbf{h}_{\alpha} = \bm{\theta}^{T} (\mathbf{h} + \alpha \bm{\theta} ) = \bar{p} + \alpha||\bm{\theta}||^{2} 
% \end{equation}%
%

By performing this intervention past all attention layers, we intervene with next token prediction and refrain from editing representation values that are cached and used in consecutive decoding steps. This way, next token predictions along the thinking trajectory are influenced in two ways: (i) locally, by the intervention-edited representation values; (ii) and historically, derived from the auto-regressive process, through the tokens that were already potentially altered as a result of previous interventions.% steps. 

In our experiments, we treat $\alpha$ as a hyperparameter that determines the strength of the intervention. Setting $\alpha = 0 $ results in no intervention, preserving the original computation. Positive values of $\alpha$ induce \textit{overclocking}. We hypothesize that overclocking will accelerate the model’s reasoning phase, making it shorter and more decisive.

\paragraph{Overthinking.}
We analyze the impact of TPV intervention on different thinking patterns. Figure~\ref{fig:teaser}(b) compares two thinking sequences generated by the DeepSeek-R1-Distill-Qwen-32B model—with and without intervention. The original sequence is marked by hesitation and verbosity, while the TPV-accelerated version is significantly more concise, using nearly six times fewer tokens. Notably, both trajectories arrive at the correct final answer. 

%\begin{wrapfigure}{r}{0.46\linewidth}
%    \vspace{-14pt}
% \begin{figure}[t]
%     \centering
%     \includegraphics[width=0.7810\linewidth]{figures/rnn_and_Zero_Shot/progress_with_without4.jpeg}
%     %\vspace{-16pt}
%     \caption{\small Predicted progress ($\bar{p}$) values for two thinking sequences, one from the original generation and the other from the TPV-intervened process, on the same problem using DeepSeek-R1-Distill-Qwen-32B for the prompt in Example~\ref{ex:ex3}. For this prompt the original generation does not converge to an answer.}
%     \label{fig:base_vs_intervention}
% \end{figure}
 %   \vspace{-10pt}
%\end{wrapfigure}

%
%
%

\begin{example}[t]
\resizebox{0.9993\textwidth}{!}{
\begin{tcolorbox}[
    colback=white,
    colframe=black,
    boxrule=0.5pt,
    sharp corners,
    boxsep=2pt,
    left=2pt,
    right=2pt,
    top=2pt,
    bottom=2pt,
]
\begin{minipage}[t]{0.495\textwidth}
\vspace{0pt}
\begin{tcolorbox}[colback=gray!5!white, colframe=black, boxrule=0.5pt, arc=3pt,title=Prompt]
\small In how many ways can 8 people sit around a round table if 3 of the people -- Pierre, Rosa, and Thomas -- all want to sit together?  (Two seatings are considered the same if one is a rotation of the other.) Please reason step by step, and put your final answer within \texttt{\textbackslash{boxed}\{\}}
\end{tcolorbox}
\vspace{-3pt}
\begin{tcolorbox}[colback=gray!10, title={TPV $\alpha=100$ (806 Thinking Tokens)}]
<think> \small Okay, so I have this problem: 8 people want to sit around a round table, $\ldots$
Alright, let's break this down. I remember that for circular permutations, the number of ways to arrange n people is (n-1)! $\ldots$
I think the method is correct because we're treating the block as a single unit, so the number of units is 6, and the circular permutations of 6 units is 5!. Then, within the block, 3! arrangements. So, 5! * 3! = 720. I think that's solid. I don't see any mistakes in the reasoning. So, the answer should be 720.
**Final Answer**
The number of ways is \textbackslash{boxed\{720\}}.
</think>

\end{tcolorbox}
\end{minipage}
\hfill
\begin{minipage}[t]{0.4985\textwidth}
\vspace{0pt}
\begin{tcolorbox}[colback=gray!10, title=Original (Continues Past 2048 Tokens)]
<think>
\small Okay, so I have this problem where I need to figure out how many ways 8 people can sit around a round table. The catch is that Pierre, Rosa, and Thomas all want to sit together. Hmm, circular permutations can be a bit tricky, but let me think through this step by step. $\ldots$ So, if I consider Pierre, Rosa, and Thomas as one block
\textbf{But wait}, $\ldots$

\textbf{But hold on}, I want to make sure I didn't miss anything. $\ldots$

\textbf{Is there another way to think about this?} Maybe by %fixing one person's position $\ldots$

\textbf{Wait}, no, that might complicate %things.
$\ldots$

\textbf{Wait}, maybe I made a mistake% in the second approach. 
$\ldots$

\textbf{Wait}, perhaps the first approach is wrong $\ldots$

\textbf{Wait}, no, I think I confused% something. Let me clarify. 
$\ldots$

\textbf{Wait}, maybe I need to think differently. $\ldots$

\textbf{Wait}, so if I fix Pierre's seat, then the block is fixed relative to Pierre, $\ldots$

\textbf{Wait}, maybe both approaches are correct, but they are counting different things. $\ldots$

\textbf{Wait}, no, $\ldots$% in circular permutations, fixing one position is a way to account for rotational symmetry.$\ldots$
\textbf{Wait}, let me think again. If we fix Pierre's seat $\ldots$

\end{tcolorbox}
\end{minipage}%
\end{tcolorbox}
\caption{Qualitative examples of indecisive loops and overthinking in the thinking process of the base model (right), which is then mitigated in the text produced by our model (bottom left).% result in indecisive Loops (right).  which mitigated show the reasoning trajectory obtained from our models,  which intervention factor of $100$
}
\label{ex:ex3}
\vspace{-17pt}
}
\end{example}

\begin{figure}[h]
    \centering
    \includegraphics[width=0.55\linewidth]{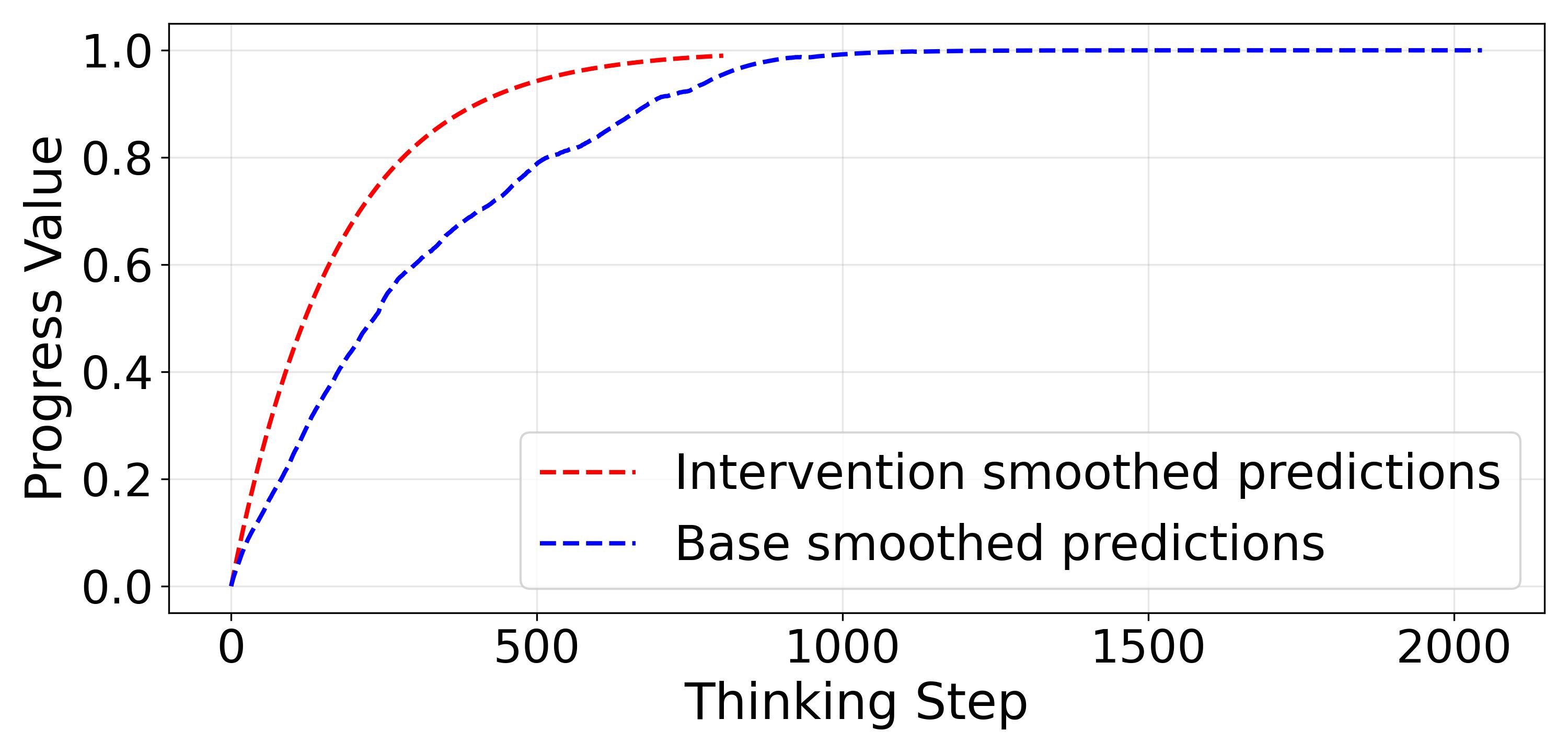}
    %\vspace{-16pt}
    \caption{Predicted progress ($\bar{p}$) values for two thinking sequences, one from the original generation and the other from the TPV-intervened process, on the same prompt in Example~\ref{ex:ex3}, using DeepSeek-R1-Distill-Qwen-32B. For this prompt the original generation does not converge to an answer.}
    \label{fig:base_vs_intervention}
\end{figure}

Example \ref{ex:ex3} illustrates a mathematical problem where the original model, when prompted, delves into a loop of self-verification and indecisiveness, raising repeating arguments and calculations without converging to a conclusion within the limit of 2048 tokens. The accelerated version reaches the correct answer while terminating in less than half of the token limit. The corresponding predicted progress trajectories for both generations are shown in Figure~\ref{fig:base_vs_intervention}. %
%The predicted progress trajectories for both generations, shown in Figure~\ref{fig:base_vs_intervention}, correspond to this example.
In addition to being shorter, the intervened trajectory shows consistently higher predicted progress % values
beyond a certain decoding step. %A sharp increase in predicted progress near the end of the accelerated sequence also aligns with the model reaching an answer.

\section{Experiments\label{sec:Experiments}}

This section presents a systematic evaluation of the intervention technique described in Sec.~\ref{subsec:MethodControl}, which overclocks the thinking phase of reasoning LLMs. We assess the method with respect to effectiveness, measured by improvements in answer quality, and efficiency, measured by lower computational cost. The experiments were conducted with two models, DeepSeek-R1-Distill-Qwen-32B and DeepSeek-R1-Distill-LLaMA-8B, which were evaluated on two mathematical benchmarks known to be effective for reasoning LLMs. The full experimental setup is detailed in Appendix~\ref{app:expirmental_Setup}.

\paragraph{Baselines.}%
To comprehensively evaluate our method, we benchmark it against the following baselines: %, each (except of the base approch) designed to accelerate the model’s thinking phase via prompting, searching or intervnation.
(i) \textbf{Base Model:} Running the model on the problem concatenated with the instruction ``Please reason step by step, and put your final answer within \texttt{\textbackslash boxed\{\}}'', as defined in DeepSeek-R1 prompting guidelines. %
(ii) \textbf{Temperature:} Same as (i) except we generate five samples and select the shortest answer. Following~\citep{guo2025deepseek}, we set the temperature to 0.6. %
% (iii) \textbf{Token-Forcing:} This prompting intervention guarantees that the model returns an answer within the compute budget, eliminating the risk of failing to provide one. We reuse the prompt from baseline (i) and intervene just before the budget limit is exceeded by halting the normal generation process and injecting the instruction: ``$\textbackslash{n}$ Based on all information gathered so far, the final answer is $\texttt{\textbackslash boxed\{\}}$". The expression inside $\verb|\boxed{}|$ is recorded as the model’s final answer. For token budgets of 256, 512, 1024, and 2048, we trigger the forcing step at tokens 250, 500, 1000, and 2000, respectively. %
(iii) \textbf{Instruct:} A straightforward prompting strategy encourages the model to produce decisive answers with the following prompt: ``Please reason step by step, place your final answer inside \verb|\boxed{}|, and then immediately stop with $<|\text{end\_of\_sentence}|>$. Present all necessary calculations or arguments concisely, avoiding unnecessary elaboration or verbosity. $<\text{think}>$''.

% \vspace{-7pt}
\subsection{Effectiveness}
% \vspace{-5pt}
We measure the effectiveness of TPVs in Tables.~\ref{tab:model1_results}, and \ref{tab:model2_results} which report results for DeepSeek-R1-Qwen-32B and DeepSeek-R1-LLaMA-8B respectively. Each table summarizes performance on two mathematical datasets evaluated under three token budgets. The right half of each table presents GSM-8K scores for budgets of 256, 512, and 1024 tokens, whereas the left half presents Math-500 scores for budgets of 512, 1024, and 2048 tokens.

For each regime we report three key metrics to evaluate model performance:
(i) \textbf{$\#$Correct} - the number of problems for which the model produced a correct final answer. We define the answer as the expression enclosed within the \texttt{\textbackslash boxed\{\}} symbol. This pattern is matched programmatically and verified manually.
(ii) \textbf{$\#$Answered} - the number of generations in which the model produced an answer, operationally determined by the presence of the \texttt{\textbackslash boxed\{\}} symbol in the output.
(iii) \textbf{$\#$Ended} - the number of generations that concluded naturally before hitting the enforced token limit.

It is important to note that the number of answered problems can exceed the number of completed generations. This is because both models often produce the \texttt{\textbackslash boxed\{\}} token at the end of the thinking phase, before completing the full response, allowing truncated generations to still include a valid answer. The opposite may also occur: particularly the LLaMA model occasionally fails to follow the instruction to wrap the answer in \texttt{\textbackslash boxed\{\}}, causing exclusion of otherwise valid answers.

Table~\ref{tab:model1_results} reveals four notable trends: %
 \textbf{(1) The impact of $\alpha$:} Increasing $\alpha$ in our method from 5 to 100, both with and without the instruction-based acceleration used in baseline (iii), generally increases the numbers of completed, ended, and correct answers produced by the model, providing further evidence that our intervention method influences the thinking length. In particular, on Math-500, larger values of $\alpha$ increased the number of correct answers by 42,39,7 on token budget of 512,1024 and 2048 respectively. Similarly, on GSM-8K, the number of ended answers increases by 15, 19, and 5 for token budgets of 256, 512, and 1024, respectively. %between the basline
%Code

%\textbf{(2) Comparisons against prompting and temperature-based acceleration techniques:}
 \textbf{(2) Comparing Acceleration Baselines Against the Base Model:} Baselines (ii) and (iii) accelerate the base model using prompting responses and temperature-based ensembling. Unsurprisingly, in most regimes both methods increase all three metrics, demonstrating that they are strong baselines against which to evaluate our overclocking approach. Specifically, on the Math-500 dataset with a 512-token budget, the number of correct answers rises from 36 (base) to 54 under the temperature baseline and 53 under the instruction-based baseline, while the number of answers increases from 38 to 59 for both methods. Finally, the number of completions increases from 28 for the base model to 43 for the temperature baseline and 52 for the instruction-based baseline. Similarly, under the same 512-token limit on GSM-8K, both acceleration strategies yield improvements. The temperature-based approach increases correct responses from 221 to 237, while the instruction-driven method produces 224 correct answers, and the total answers climb from 223 to 249 with the temperature strategy and to 237 with instruction prompts. Completions also see a marked rise, increasing from 28 for the base model to 43 under the temperature baseline and to 52 with instruction-based acceleration. %

\textbf{(3) Comparing Our Method Against Baselines:} Although these baselines perform strongly and the temperature-based baseline requires roughly five times more compute, our method outperforms both by producing more correct answers and more decisive responses. In particular, under low compute budgets of 256 or 512 tokens, our approach with $\alpha = 100$ increases the number of correct answers on Math-500 by at least 80\% in the 512 token-budget regime and boosts correct responses on GSM-8K by an average of 80\% across the 256 and 512 token settings. For example, on Math-500 in the 512 token-budget regime, correct answers rise from 54 under the temperature baseline and 53 under the instruction baseline to 100 with our method, resulting in a gain of more than 46 correct answers, while the number of completed answers increases from 59 under both baselines to 106, a gain of 47. Remarkably, these increases in correct answers do not come at the cost of more errors, as the error rate remains unchanged. This suggests our method shortens reasoning without increasing errors, promoting more decisive thinking. Our findings for compute budgets larger than 512 generally follow the same trend, showing improvements in the number of correct answers in most regimes, without an increase in the error rate.

 \textbf{(4) Complementary Contribution:} Although our empirical findings approve that our method is more effective than the baselines, there are still cases where our method lags behind the prompting-based approach (denoted as 'Instruct'). One prominent example is the regime of token-budget of 2048 on the Math 500, where the instruct baseline answers correctly 10\% more than our method. A similar trend was also observed in Table~\ref{tab:model2_results}, with experiments conducted using LLaMA instead of the Qwen DeepSeek variants. This observation raises the question of whether the improvements are orthogonal or if the two methods directly compete. To investigate, we conducted dedicated experiments in which the instruction-based prompting technique was combined with our intervention method and compared against each individual approach. As shown in the last two rows of both tables, this hybrid approach consistently yields the best performance across most regimes, achieving an improvement of $66\%$ on average and {$285\%$} at max over the instruction-based baseline, and {$223\%$} on average and {$1416\%$} at max over the base model. These findings suggest that our method is complementary to prompting strategies and can be effectively integrated with other acceleration techniques.

\begin{table*}[t]
  \caption{\textbf{Results for DeepSeek-R1-Qwen-32B} on GSM-8K and Math500 at different context lengths (512, 1024, 2048 for Math500; 256, 512, 1024 for GSM-8K), showing \#Correct '\#Cr', \#Answered '\#An' and \#Ended '\#En' for each.}
  \vspace{-4pt}
  \label{tab:model1_results}
  {
  \begin{center}
  \resizebox{0.995\textwidth}{!}{
    \begin{tabular}{@{}l@{~~~}
      ccc@{~~}ccc@{~~}ccc@{~~~}
      ccc@{~~}ccc@{~~}ccc@{}}
    \toprule
    \multirow{3}{*}{\textbf{Method}}
      & \multicolumn{9}{c}{\textbf{Math500}}
      & \multicolumn{9}{c}{\textbf{GSM-8K}} \\
    \cmidrule(lr){2-10} \cmidrule(lr){11-19}
      & \multicolumn{3}{c}{\textbf{512}}
      & \multicolumn{3}{c}{\textbf{1024}}
      & \multicolumn{3}{c}{\textbf{2048}}
      & \multicolumn{3}{c}{\textbf{256}}
      & \multicolumn{3}{c}{\textbf{512}}
      & \multicolumn{3}{c}{\textbf{1024}} \\
    \cmidrule(lr){2-4} \cmidrule(lr){5-7} \cmidrule(lr){8-10}
    \cmidrule(lr){11-13} \cmidrule(lr){14-16} \cmidrule(lr){17-19}
      & \#Cr & \#An & \#En
      & \#Cr & \#An & \#En
      & \#Cr & \#An & \#En
      & \#Cr & \#An & \#En
      & \#Cr & \#An & \#En
      & \#Cr & \#An & \#En \\
    \midrule
    Base          & 36  & 38  & 28  & 154  & 156  & 101  & 278  & 280  & 245  
                  & 12  & 16  & 9   & 221  & 223  & 227  & 285  & 298  & 296 \\
    Temperature   & 54  & 59  & 43  & 146  & 149  & 119  & 294   & 305   & 294  
                  & 39  & 40  & 33  & 237  & 249  & 248  & 237  & 249  & 300 \\
    % Token Forcing & 201  & N.A  & 172  & 266   & N.A   & 216   & 357   & N.A   & 316    
    %               & 255  & N.A  & 195  & 271   & N.A   & 289   & 285   & N.A   & 299 \\
    Instruct      & 53  & 59  & 52  & 211  & 214  & 177  & \textbf{316}  & 321  & 296  
                  & 16  & 18  & 15  & 224  & 237  & 232  & 285  & 299  & 300 \\
    \midrule
    TPV \(\alpha=5\) & 58   & 58   & 42  & 162   & 165   & 98  & 277  & 280  & 257    
                  & 23  & 26  & 14  & 226  & 237  & 233  & 285  & 297  & 293 \\
    TPV \(\alpha=100\) & 100  & 106  & 94  & 201  & 213  & 167  & 284  & 296  & 268  
                  & 36  & 41  & 29  & 245  & 256  & 252  & 285  & 299  & 298 \\
    TPV \(\alpha=5\) Ins & 123  & 126  & 109 & 240   & 254   & 218 & \textbf{316}  & 331  & 306    
                  & 34  & 26  & 34  & 239  & 250  & 248  & \textbf{286}  & 300  & 300 \\
    TPV \(\alpha=100\) Ins & \textbf{148}  & 160  & 142 & \textbf{256}  & 290  & 279  & 300  & 337  & 330   
                  & \textbf{45}  & 49  & 40  & \textbf{247}  & 263  & 261  & 284  & 298  & 298 \\
    \bottomrule
    \end{tabular}
  }
  \end{center}
  }

\end{table*}

\begin{table*}[t]
  \caption{\textbf{Results for DeepSeek-R1-LLaMA-8B} on GSM-8K and Math500 at different context lengths (512, 1024, 2048 for Math500; 256, 512, 1024 for GSM-8K), showing \#Correct '\#Cr', \#Answered '\#An' and \#Ended '\#En' for each.}
 \vspace{-4pt}
  \label{tab:model2_results}
  \begin{center}
  \resizebox{0.995\textwidth}{!}{
    \begin{tabular}{@{}l@{~~~}
      ccc@{~~}ccc@{~~}ccc@{~~~}
      ccc@{~~}ccc@{~~}ccc@{}}
    \toprule
    \multirow{3}{*}{\textbf{Method}}
      & \multicolumn{9}{c}{\textbf{Math500}}
      & \multicolumn{9}{c}{\textbf{GSM-8K}} \\
    \cmidrule(lr){2-10} \cmidrule(lr){11-19}
      & \multicolumn{3}{c}{\textbf{512}}
      & \multicolumn{3}{c}{\textbf{1024}}
      & \multicolumn{3}{c}{\textbf{2048}}
      & \multicolumn{3}{c}{\textbf{256}}
      & \multicolumn{3}{c}{\textbf{512}}
      & \multicolumn{3}{c}{\textbf{1024}} \\
    \cmidrule(lr){2-4} \cmidrule(lr){5-7} \cmidrule(lr){8-10}
    \cmidrule(lr){11-13} \cmidrule(lr){14-16} \cmidrule(lr){17-19}
      & \#Cr & \#An & \#En
      & \#Cr & \#An & \#En
      & \#Cr & \#An & \#En
      & \#Cr & \#An & \#En
      & \#Cr & \#An & \#En
      & \#Cr & \#An & \#En \\
    \midrule
    Base       &  6  &  6   &  11   &  47   &  54   &  124  & 112 &  122  &  273 &  0   &  0   & 2  & 21    &  27   &  99  & 40 & 47 & 247  \\
    Temperature     &  29   & 31  & 51 & 95 & 104 & 229 & 137 & 156 & 332 & 7 & 7 & 20 & 50 & 60 & 216 & 63 & 74 & 294 \\
    % Token Forcing & --  & --  & N.A  & --   & --   & N.A   & --   & --   & N.A    
    %               & --  & --  & N.A  & --   & --   & N.A   & --   & --   & N.A \\
    Instruct        &  50  & 55   &  68  &  107   & 117  & 190  &  153   & 172    & 302  &  7   & 8 & 13    &  66 & 81 &  167 &  \textbf{85} & 100 &  273 \\
    \midrule
    TPV \(\alpha=5\)          & 4   & 5   & 12  & 37  & 42  & 118 & 109 & 117 & 276 & 0   & 0   & 2   & 22  & 26  & 95  & 35  & 43  & 235 \\
    TPV \(\alpha=100\)       & 26  & 30  & 41  & 70  & 77  & 170 & 102 & 113 & 273 & 4   & 5   & 7   & 28  & 34  & 118 & 38  & 46  & 223 \\
    TPV \(\alpha=5\) Ins       & 51  & 55  & 67  & 103 & 112 & 193 & \textbf{159} & 175 & 315 & 12  & 12  & 13  & 58  & 72  & 164 & 80  & 95  & 268 \\
    TPV \(\alpha=100\) Ins     & \textbf{91}  & 118 & 119 & \textbf{124} & 170 & 227 & 142 & 192 & 284 & \textbf{27}  & 33  & 70  & \textbf{80}  & 96  & 176 & 81  & 104 & 211 \\
    \bottomrule
    \end{tabular}
  }
  \end{center}
  % \vspace{-9pt}
\end{table*}

% \vspace{-8pt}
\subsection{Efficiency}
\begin{figure}[t]
\centering
    \includegraphics[width=.7\linewidth]{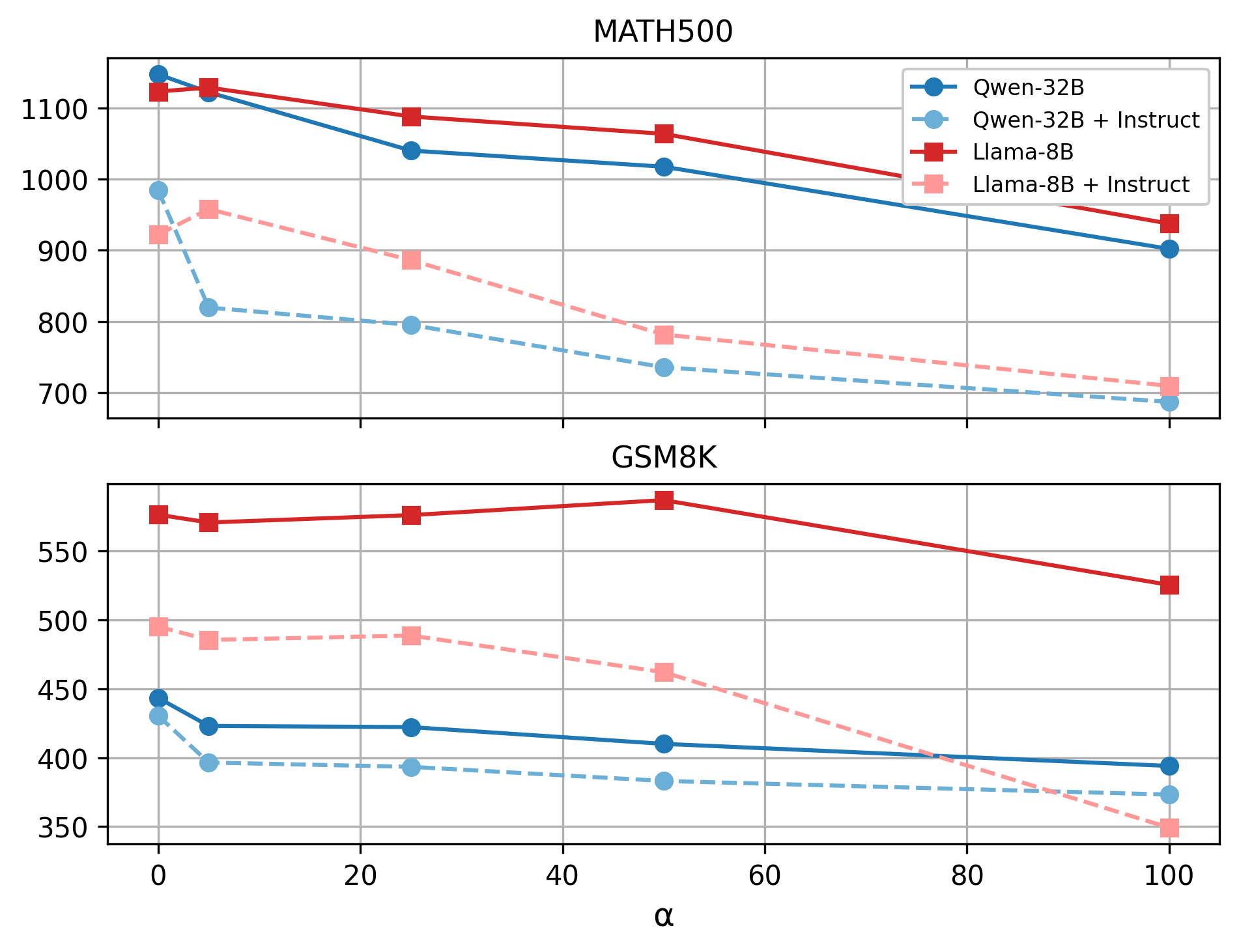}
  %  \vspace{-17pt}
    \caption{\textbf{Impact of $\alpha$ on Intervention:} The x-axis represents different values of $\alpha$ in $\{0, 5, 25, 50, 100\}$, where $\alpha = 0$ indicates no intervention and higher values of $\alpha$ %increase the value of the TPV
    amplify the amount of intervention. The y-axis shows the average number of tokens required to complete the answer. Results are obtained using DeepSeek-R1 models.}
    % PlaceHolder - TD LLaMA results \lw{NEED TO HAVE THE X AXIS IN BOTH PLOTS NEED TO WRITE ALPHA AS THE GREEK LETTER NEED TO EXPLINA WHAT IS THE Y VALUE ETC}}
 %   \vspace{-14pt}
\label{fig:Intervention}
\end{figure}

% \vspace{-5pt}
%
We conduct a series of intervention experiments on the Math-500 and GSM8K datasets by varying the intervention parameter $\alpha$ to overclock the model’s thinking phase. Figure~\ref{fig:Intervention} shows that increasing $\alpha$ consistently reduces the length of the thinking phase, making the reasoning process more efficient. These findings support our hypothesis that TPV functions as an active control signal in the model’s internal computation, rather than being a passive correlate. In particular, when applying our method with $\alpha = 100$ to the DeepSeek-R1 LLaMA model using the prompting strategy (baseline iii) on the GSM8K dataset, the average number of tokens decreases from approximately 500 to fewer than 350, resulting in a 30\% reduction in compute. Moreover, all positive values of $\alpha$ consistently accelerate the thinking phase relative to the baseline ($\alpha = 0$), while also improving its effectiveness, as shown in Tables~\ref{tab:model1_results}--\ref{tab:model2_results}. For additional qualitative examples, see examples in Appendix~\ref{sec:AdditionalQualitativeAnalysis}.

%\begin{wrapfigure}{r}{0.51\linewidth}
%\begin{figure}
%    \vspace{-39pt}
%\end{wrapfigure}
% \end{figure}

%

%with the, we observe that setting a highly aggressive intervention value of $\alpha=100$ reduces the thinking length by more than 50\%, from over 1100 tokens to approximately 550 \lw{NNOT SHOWN IN THE FIGURE WHICH STOPS AT 100-Watinig for final figure}.

% \vspace{-6pt}
\subsection{Qualitative Analysis}
\begin{figure}[t]
    % \vspace{-5pt}
    \includegraphics[width=0.5\linewidth]{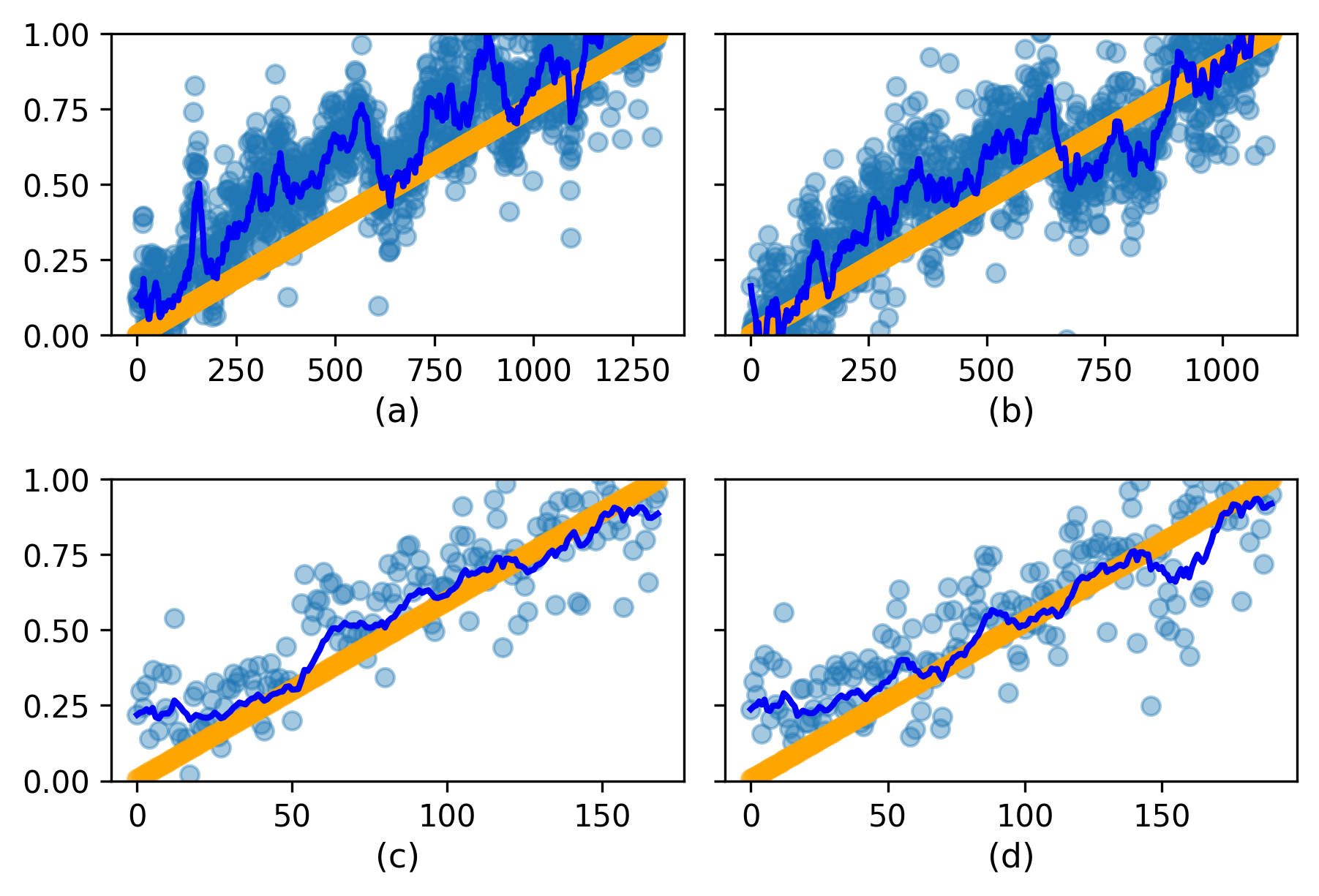}
    \includegraphics[width=0.48\linewidth]{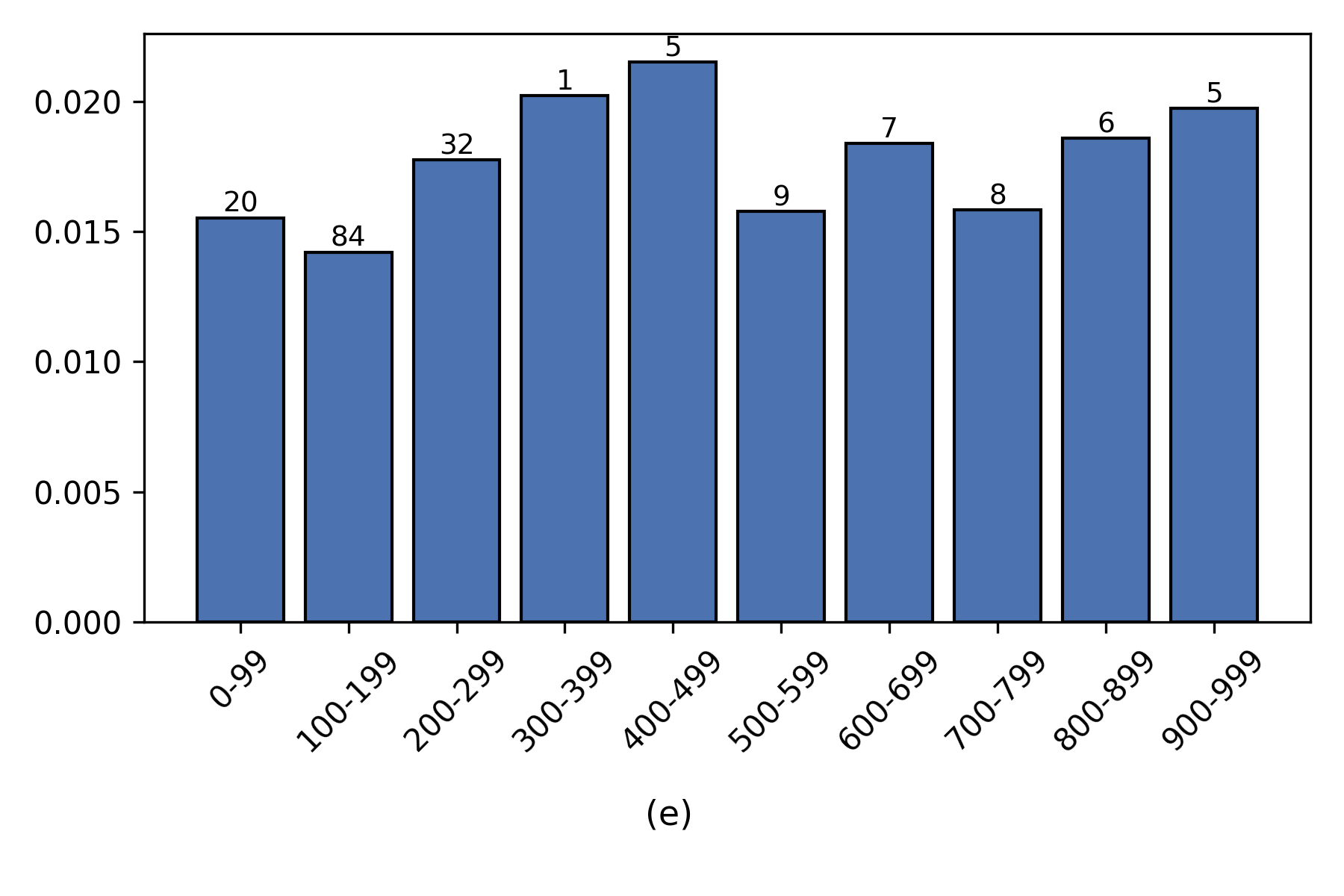}
    % \includegraphics[width=0.33\linewidth]{figures/loading_bar_regression.png}
    % \vspace{-7pt}
    \caption{\textbf{Robustness of predicted progress under varied prompting styles and reasoning lengths.}
\textbf{Left:} Predicted and smoothed progress trajectories for the problem “How many 3-digit numbers can be formed using the digits 1 through 9 (no zeros), with no repeated digits?” under four different prompts:
(a) “Please reason step by step.”
(b) “Think carefully and slowly. Provide a detailed explanation.”
(c) “Think quickly and provide a concise answer.”
(d) “Present all necessary calculations or arguments concisely; avoid unnecessary elaboration or verbosity.”
\textbf{Right:} (e) Mean MSE test loss from both Math500 and GSM8K datasets, grouped by thinking phase length (binned). The number of trajectories in each bin is shown above each bar.}  
% \vspace{-9pt}
\label{fig:reliabilityProgressVectors}
\end{figure}

% \vspace{-2pt}
To further evaluate the reliability of TPVs in estimating a model’s position within its reasoning process, we test their performance under two additional conditions: (i) varied prompting strategies and (ii) differing reasoning sequence lengths. Figures~\ref{fig:reliabilityProgressVectors}(a-d) show that TPVs remain effective across diverse instructions, distinct from the original prompt used during training (see caption for the prompts).  Figure~\ref{fig:reliabilityProgressVectors}(e) shows consistently low test loss across bins of different thinking sequence lengths, indicating robustness to varying reasoning depths.

To better understand how progress is monitored and how it is modified during the thinking phase, we explored cases in which the behavior of the indicator is not monotonic, see Appendix~\ref{app:nonmonotonic}. For example it is often the case  that  
the progress prediction increases, until a noticeable drop occurs. 
This drop in predicted progress often coincides with self-verification behavior, where the model attempts to solve the problem again using an alternative approach.

To further explore TPV’s capacity to capture and explain different reasoning dynamics, we examine how predicted progress responds to specific tokens that suggest either hesitation (``wait'') or advancement (``right''). We quantify this behavior, by measuring the average effect of individual tokens on predicted progress values across the dataset, see Appendix~\ref{app:wordeffect}. The analysis shows that certain words do tend to significantly shift the predicted progress indicator in a characteristic direction.

% \vspace{-5pt}
\section{Discussion\label{sec:Discussion}}
% \vspace{-7pt}
%\paragraph{Linking Empirical Findings to Metacognitive Theory}
%Our empirical findings show that reasoning models such as DeepSeek-R1 possess a mechanism for tracking and controlling their relative progress during the thinking phase. We argue that this opens a unique opportunity to examine the cognitive-like abilities of such models. %%To support this claim, we first recall that in Section~\ref{sec:method} we showed that LLMs track their progress using progress vectors, which are also employed to control the desired length of the thinking phase. We now turn to key concepts from metacognition and examine how they are associated with our empirical findings.

The Nelson and Narens metacognition model~\citep{nelson1990metamemory} distinguishes two interacting levels: the object level, which performs cognitive operations, and the meta level, which engages in monitoring and control over the object-level processes. We show that these two information flows, monitoring and control, are manifested in LLMs, as described in Sections~\ref{subsec:MethodMonitoring}, and \ref{subsec:MethodControl}, respectively. %Specifically, by examining the progress vectors through the lens of the progress probe, we show that the thinking tokens generated by the model influence its progress vector, indicating a form of monitoring. Moreover, our intervention experiments demonstrate that, based on this tracking, the model adjusts and controls the length of the thinking phase, as shown in Figure~\ref{fig:Intervention}.
Similarly, \citet{zimmerman2002becoming} describe the process of self-regulated learning and highlights monitoring and self-reflection as two essential components. In this context, our work shows that LLMs can estimate the number of reasoning steps required to reach a desired answer, and that this estimation evolves in real time based on the tokens generated. %These findings indicate that the model engages in both monitoring and iterative self-reflection during the reasoning process.

%\vspace{-3pt}
%\paragraph{Enhancing Human-AI Collaboration Through Transparent and Controllable Reasoning}
On the applicative side, our method represents an important step toward making reasoning models more transparent and controllable. %, thereby enabling smoother and more effective collaboration with human users. 
At the time of writing, millions of users interact with reasoning models such as OpenAI’s O1 or DeepSeek-R1. During inference, particularly in the ``thinking phase,'' users are often left without visibility into the model’s internal progress or how close it is to producing an answer. By introducing an interactive progress bar that visualizes the model’s internal status during reasoning, we make these models more predictable, responsive, and user-friendly. %This allows users to better understand where the model may be struggling, which reasoning steps are difficult, and how their prompts might be refined.
Furthermore, our approach introduces a plug-and-play intervention mechanism that enables real-time acceleration of the reasoning process through overclocking. This feature is especially useful for users who have an intuition about the complexity of their task and wish to adapt the model’s reasoning depth accordingly. %{\color{red}for example, ..online} By adjusting the length of the thinking phase based on task difficulty or real-time constraints, our method improves efficiency while maintaining or even enhancing performance. More broadly, 
This contributes to the ongoing effort to make reasoning-capable language models more interactive, adaptive, and aligned with human needs. %As such, we consider this work a promising step toward advancing collaborative reasoning between humans and AI systems.

\paragraph{Limitations.}
% \vspace{-3pt}
While our approach demonstrates promising results for monitoring and controlling thinking path lengths in LLMs, several important limitations should be acknowledged. 
First, our investigations are primarily focused on mathematical reasoning tasks, which represent a well-structured domain with clear evaluation metrics. The applicability of our progress vector technique to more open-ended reasoning domains (e.g., ethical reasoning, creative problem-solving, or multi-step planning) remains unexplored. 
Specifically, while overclocking improves performance on our tested benchmarks,  we have not thoroughly investigated potential negative side effects of progress vector manipulation on reasoning coherence or factual accuracy outside our test distributions.
Second, our method requires access to model hidden states during inference, making it challenging to implement in black-box or API-only settings where only input-output functionality is available. %This limits immediate practical applications to scenarios where researchers or developers have full access to model internals.
Finally, our experiments focus exclusively on the length dimension of reasoning, without addressing other important aspects such as reasoning breadth, diversity of approaches considered, or uncertainty estimation.

% \vspace{-6pt}
\section{Conclusions}
% \vspace{-5pt}

This paper introduces a novel approach to monitoring and controlling thinking path lengths, revealing fundamental mechanisms by which LLMs regulate their reasoning processes. %Our investigation demonstrates that reasoning models encode a form of progress awareness in their hidden representations, enabling us to extract this signal through progress vector projections. This discovery provides empirical evidence that LLMs maintain an internal representation of their reasoning trajectory—a capability previously associated primarily with cognitive systems. 
By identifying and manipulating the progress mechanism, we have developed two practical applications with significant implications for AI systems: (i) a dynamic thinking progress visualization that enhances transparency, and (ii) an ``overclocking'' intervention technique that significantly improves both efficiency and performance. %Our empirical results demonstrate that overclocking can reduce thinking length by approximately 50\% while simultaneously improving performance by up to 30\% on mathematical reasoning benchmarks like Math-500.
The ability to modulate reasoning depth without sacrificing—and often enhancing—performance challenges conventional assumptions about the relationship between computation and accuracy in LLMs. This suggests that modern LLMs are capable of more efficient reasoning but may benefit from external guidance to avoid overthinking patterns. %Beyond the technical contributions, this work bridges interpretability research with metacognitive theory, offering insights into how artificial systems might implement forms of self-monitoring and self-regulation.

\section{Acknowledgments}
This work was supported by a grant from the Tel Aviv University Center for AI and Data Science (TAD). This research was also supported by the Ministry of Innovation, Science \& Technology ,Israel (1001576154) and the Michael J. Fox
Foundation (MJFF-022407).

\bibliographystyle{plainnat} % or unsrtnat, abbrvnat … pick one and stay consistent
\bibliography{references} 

%%%%%%%%%%%%%%%%%%%%%%%%%%%%%%%%%%%%%%%%%%%%%%%%%%%%%%%%%%%%
\newpage
\appendix

\section{Experimental Setup\label{app:expirmental_Setup}}
This section describes the experimental setup used for the empirical studies in Sec.~\ref{sec:Experiments}. All experiments were conducted in PyTorch with official Hugging Face models and DeepSeek prompts. Each run used a single A100 GPU and finished within at most two days. 

\paragraph{Datasets.}%
We evaluated performance on two publicly available mathematics datasets: GSM-8K and Math-500. For Math-500, we randomly sampled 80 problems to train our loading-bar regressor and used the remaining 420 problems as the test set. For GSM-8K, we randomly sampled 330 problems, reserving the first 30 for training and the remaining 300 for testing. These datasets were chosen because they effectively assess the reasoning capabilities of LLMs, as reasoning has been shown to be remarkably effective on mathematical problems.

\section{Non-monotonic Progress Prediction\label{app:nonmonotonic}}
\begin{figure}[h]
    \centering
            \includegraphics[width=0.79\linewidth]{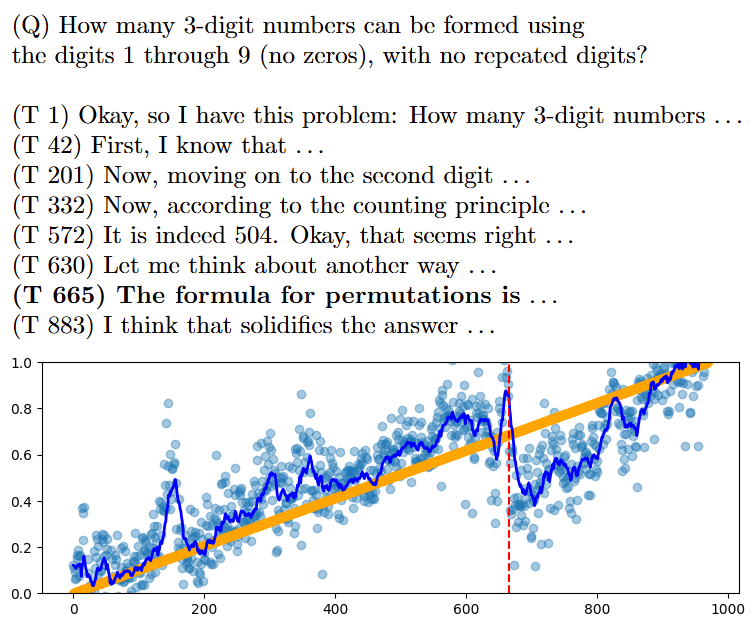}\\
    \caption{A math question and the associated thinking sequence generated by DeepSeek-R1-Distill-Qwen-32B, with each row indicating the index of its first token. The sequence is paired with its TPV-predicted progress trajectory, highlighting the effect of self-verification behavior within the model’s reasoning process. A vertical line at token index 665 marks a notable drop in predicted progress, aligning with the point where the model begins re-solving the problem.
}
    \label{fig:monotonic}
\end{figure}

We perform a qualitative analysis of TPV-predicted progress trajectories and find that non-monotonic patterns often correspond to self-verification behaviors in the model’s reasoning. In such cases, the model reapproaches the problem using a different method to confirm its solution. For instance, as illustrated in Figure~\ref{fig:monotonic}, the model first solves a combinatorics question using the counting principle, then revalidates its answer by applying the permutations formula. This second solving attempt begins at token 665—indicated by a vertical line—which coincides with a clear drop in the predicted progress, reflecting the model’s re-evaluation phase.

\section{Effect of Specific Words on Predicted Progress \label{app:wordeffect}}

When examining thinking sequences qualitatively, certain tokens frequently appear in contexts that suggest either hesitation or progression toward concluding the reasoning process. Words such as “hmm,” “wait,” and “right” often signal meaningful transitions in the model’s internal state. Some of these tokens reflect a recognition of inconsistency or error, prompting the model to adjust its reasoning strategy. Others indicate the completion of a reasoning step and a shift toward the final solution. TPVs enable a principled way to quantify this behavior.

\begin{figure}
     \centering
            \includegraphics[width=0.755\linewidth]{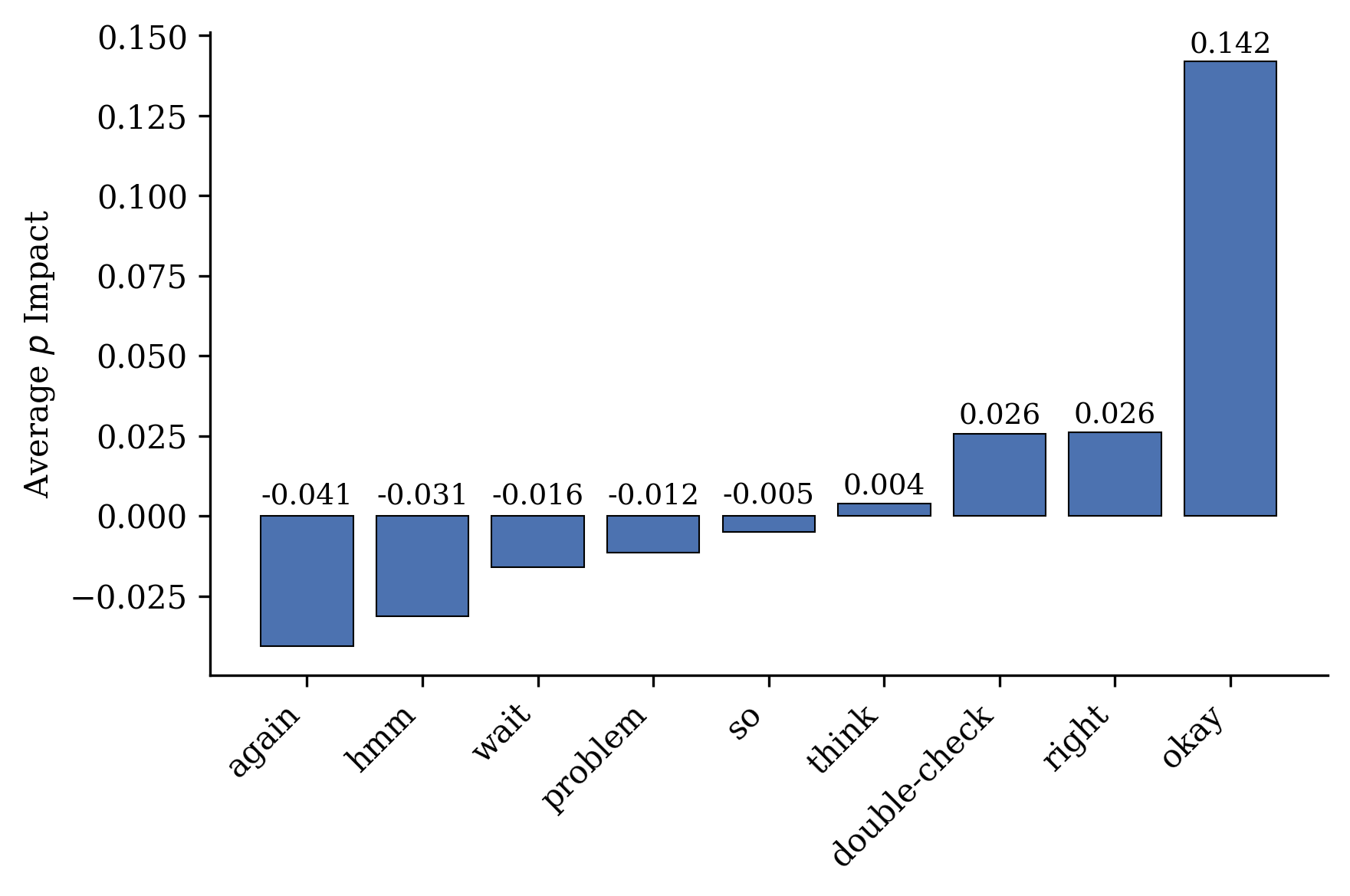}
    \caption{Average change in $\bar{p}$ values caused by significant tokens across the dataset, highlighting their impact on predicted progress.}
\label{fig:perword}
\end{figure}

 We analyze the average influence of individual tokens on predicted progress values across the dataset. As illustrated in Figure~\ref{fig:perword}, tokens with the most negative impact—such as “again,” “hmm,” “wait,” and “problem”—are closely tied to hesitation or reevaluation. In contrast, tokens like “right,” “so,” and “okay” are associated with upward shifts in predicted progress, signaling growing confidence and momentum in the model’s reasoning process.

% \begin{wrapfigure}{hr}{0.5\linewidth}

%%%%%%%%%%%%%%%%%%%%%%%%%%%%%%%%%%%%%%%%%%%%%%%%%%%%%%%%%%%%
\section{Additional Qualitative Analysis\label{sec:AdditionalQualitativeAnalysis}}
To highlight the impact of our intervention in mitigating overthinking, we present a qualitative analysis in Examples~\ref{ex:ex2}--\ref{ex:exEquations}. Examples~\ref{ex:exFibonacci}--\ref{ex:exEquations} demonstrate that while both the baseline model and its over-clocked variant arrive at the same answer, the over-clocked model does so more decisively and with significantly fewer thinking tokens. In contrast, Example~\ref{ex:ex2} shows a case where the base model does not succeed in providing an answer within the token limit, in contrast to the over-clocked variant, which provides an answer after 259 thinking tokens.

\begin{example}[t]
\resizebox{0.9993\textwidth}{!}{
\begin{tcolorbox}[
    colback=white,
    colframe=black,
    boxrule=0.5pt,
    sharp corners,
    boxsep=2pt,
    left=2pt,
    right=2pt,
    top=2pt,
    bottom=2pt,
]
\begin{minipage}[t]{0.495\textwidth}
\vspace{0pt}
\begin{tcolorbox}[colback=gray!5!white, colframe=black, boxrule=0.5pt, arc=3pt,title=Prompt]
\small Suzanne walks four miles every third day. What is the fewest number of miles she can walk in February? Please reason step by step, and put your final answer within \texttt{\textbackslash{boxed}\{\}}
\end{tcolorbox}
\vspace{-3pt}
\begin{tcolorbox}[colback=gray!10, title={TPV $\alpha=100$ (259 Thinking Tokens)}]
<think> \small Suzanne walks 4 miles every 3 days. To find the fewest number of miles she can walk in February, we need to determine how many 3-day periods are in 28 days (a non-leap year February).

1. **Calculate the number of 3-day periods in 28 days:**
   \[
   \frac{28}{3} \approx 9.33
   \]
   Since Suzanne can't walk a fraction of a 3-day period, we take the integer part, which is 9.

2. **Determine the total miles walked:**
   \[
   9 \times 4 = 36 \text{ miles}
   \]

3. **Check if 28 days is a multiple of 3:**
   \[
   28 \div 3 = 9 \text{ with a remainder of 1}
   \]
   Since 28 is not a multiple of 3, 9 periods of 3 days account for 27 days, leaving 1 day. Suzanne doesn't walk on the 28th day, so 36 miles is the minimum.

**Final Answer**
The number of ways is \textbackslash{boxed\{36\}}.
</think>

\end{tcolorbox}
\end{minipage}
\hfill
\begin{minipage}[t]{0.4985\textwidth}
\vspace{0pt}
\begin{tcolorbox}[colback=gray!10, title=Original (Continues past 2048 tokens)]
<think>
\small Okay, so I need to figure out the fewest number of miles \ldots

First, I should know how many days are in February. \ldots maybe I can assume it's a non-leap year with 28 days? Or does it matter? Let me think. \ldots

Let me list the days she walks:

Day 1, 4, 7, 10, 13, 16, 19, 22, 25, 28.

\textbf{Wait}, that's 10 walks. \ldots
Maybe she can start on a later day to minimize the number of walks in February. \ldots

% Ah, that's a good point. If she starts walking on a day later in the month, she might have fewer walks. So, to minimize the number of walks, she should start as late as possible.

But how late can she start? \ldots if she starts on day 26, for example, she would walk on day 26, then day 29, but February only has 28 days, \ldots she would only walk once in February. \ldots

\textbf{Wait}, actually, the problem says she walks four miles every third day. \ldots

% I think it's the former: she walks every third day, meaning she has a consistent schedule where she walks every three days. So, if she starts on day 1, she walks on day 1, 4, 7, etc. If she starts on day 2, she walks on day 2, 5, 8, etc. So, the starting day affects how many walks she has in February.

% Therefore, to minimize the number of walks, she should start as late as possible. So, starting on day 26, she walks once on day 26. Starting on day 27, once on day 27. Starting on day 28, once on day 28.

\textbf{But wait}, if she starts on day 26, does that mean she didn't walk on day 23? \ldots she should have walked on day 23 as well. \ldots

% Hmm, this is a bit confusing. Let me clarify.

If she walks every third day, that means her walking days are spaced three days apart. So, if she walks on day x, she must have walked on day x - 3, x - 6, etc., and will walk on day x + 3, x + 6, etc. \ldots

% Therefore, if she walks on day 26, she must have walked on day 23, 20, 17, 14, 11, 8, 5, and 2. So, starting from day 2, she walks every third day.

% But if we're only considering February, which has 28 days, starting from day 2, she would walk on day 2, 5, 8, 11, 14, 17, 20, 23, 26, and 29. But day 29 is beyond February, so in February, she walks 9 times.

% Wait, that's more than the 10 walks if she started on day 1. Wait, no, starting on day 1, she walks on day 1, 4, 7, 10, 13, 16, 19, 22, 25, 28, which is 10 walks.

% Starting on day 2, she walks on day 2, 5, 8, 11, 14, 17, 20, 23, 26, which is 9 walks in February.

% Similarly, starting on day 3, she walks on day 3, 6, 9, 12, 15, 18, 21, 24, 27, which is 9 walks.

\textbf{Wait}, so starting on day 2 or 3, she has 9 walks, whereas starting on day 1, she has 10 walks.

% So, to minimize the number of walks, she should start on day 2 or 3, resulting in 9 walks, which is 36 miles.

\textbf{But earlier}, I thought she could start on day 26 and only walk once, but that seems incorrect because her walking schedule is every third day, so she can't just start on day 26 without having walked on day 23, 20, etc. \ldots

% Therefore, the number of walks depends on the starting day, but the starting day affects the total number of walks in February.

% So, to find the fewest number of miles, we need to find the starting day that results in the fewest number of walks in February.

% Let me think about this systematically.

% February has 28 days. Let's number them from 1 to 28.

% If Suzanne walks every third day, her walking days can be represented as an arithmetic sequence with a common difference of 3.

% The number of terms in this sequence within 1 to 28 will determine the number of walks.

% To minimize the number of terms, we need to choose the starting day such that the sequence includes as few terms as possible \ldots

% Starting on day 27, next walk is day 30, which is beyond February.

% Starting on day  $\ldots$

\end{tcolorbox}
\end{minipage}%
\end{tcolorbox}
\caption{An example of overthinking through first presenting an argument not aligned with the constraints of the problem and the effort of reconciling it (right); and a straightforward answer presenting a simple solution (left).% result in indecisive Loops (right).  which mitigated show the reasoning trajectory obtained from our models,  which intervention factor of $100$
}
\label{ex:ex2}
\vspace{-21pt}
}
\end{example}

\begin{example}[t]
\resizebox{0.9993\textwidth}{!}{
\begin{tcolorbox}[
    colback=white,
    colframe=black,
    boxrule=0.5pt,
    sharp corners,
    boxsep=2pt,
    left=2pt,
    right=2pt,
    top=2pt,
    bottom=2pt,
]
\begin{minipage}[t]{0.495\textwidth}
\vspace{0pt}
\begin{tcolorbox}[colback=gray!5!white, colframe=black, boxrule=0.5pt, arc=3pt,title=Prompt]
\small Compute the 9th Fibonacci number, assuming F(1) = 1. Please reason step by step, and put your final answer within \texttt{\textbackslash{boxed}\{\}}
\end{tcolorbox}
\vspace{-3pt}
\begin{tcolorbox}[colback=gray!10, title={TPV $\alpha=100$ (293 Thinking Tokens)}]
<think> \small To compute the 9th Fibonacci number, I start by understanding the Fibonacci sequence. The sequence is defined such that each number is the sum of the two preceding ones, with the first two numbers being 1.

I will list out the Fibonacci numbers up to the 9th term. Starting with F(1) = 1 and F(2) = 1, I can calculate each subsequent term by adding the two previous terms:

F(3) = F(2) + F(1) = 1 + 1 = 2

\ldots

F(9) = F(8) + F(7) = 21 + 13 = 34

By following this sequence, I find that the 9th Fibonacci number is 34.
</think>

\end{tcolorbox}
\end{minipage}
\hfill
\begin{minipage}[t]{0.4985\textwidth}
\vspace{0pt}
\begin{tcolorbox}[colback=gray!10, title=Original (1872 Thinking Tokens)]
<think>
\small Okay, so I need to find the 9th Fibonacci number, and they mentioned that F(1) is 1. Hmm, Fibonacci numbers\ldots

I remember they start with 0 and 1, but \ldots

The Fibonacci sequence is where each number is the sum of the two preceding ones. So, if F(1) is 1, then what is F(2)? I think it's also 1 because usually, the sequence starts with 0, 1, 1, 2, 3, 5, etc. \ldots

Let me write down the sequence step by step to make sure I get it right.

F(1) = 1

F(2) = 1

\ldots

F(9) = F(8) + F(7) = 21 + 13 = 34

\textbf{Wait, so F(9) is 34?}

\textbf{Let me double-check} my calculations to make sure I didn't make a mistake.

Starting from F(1) = 1:

1. F(1) = 1

2. F(2) = 1

\ldots

9. F(9) = 21 + 13 = 34

Yes, that seems correct. Each time, I'm adding the two previous numbers to get the next one. So, the 9th Fibonacci number is 34. \ldots

\textbf{let me recall another way} \ldots 

I remember there's Binet's formula, which uses the golden ratio. The formula is:

$$ F(n) = (\Phi^n - \Psi^n) / \sqrt{5} \ldots $$

Let me try using this formula to compute F(9) \ldots

First, compute \ldots

Now, compute \ldots

Let me compute this step by step: \ldots

\ldots

Similarly, compute \ldots

\ldots

Now, plug these into Binet's formula:

$$ F(9) = (\Phi^9 - \Psi^9) / \sqrt{5} \approx$$
$$(76.013 - (-0.01316)) / 2.23607 \approx \ldots \approx 34.0 $$

Wow, that's exactly 34. So, using Binet's formula also gives me 34 for F(9). That confirms my earlier result.

\textbf{Alternatively}, I could have used the recursive formula \ldots

\textbf{Just to make sure}, let me list out the Fibonacci numbers from F(1) to F(9):

1. F(1) = 1

2. F(2) = 1

\ldots

9. F(9) = 34

Yep, that's consistent. So, I'm confident that the 9th Fibonacci number is 34.
</think>

\end{tcolorbox}
\end{minipage}%
\end{tcolorbox}
\caption{An example of overthinking through excessive self-verification and redundant, duplicated computation in the thinking process of the base model (right), which is mitigated in the text produced by our model (bottom left).% result in indecisive Loops (right).  which mitigated show the reasoning trajectory obtained from our models,  which intervention factor of $100$
}
\label{ex:exFibonacci}
\vspace{-21pt}
}
\end{example}

\begin{example}[t]
\resizebox{0.9993\textwidth}{!}{
\begin{tcolorbox}[
    colback=white,
    colframe=black,
    boxrule=0.5pt,
    sharp corners,
    boxsep=2pt,
    left=2pt,
    right=2pt,
    top=2pt,
    bottom=2pt,
]
\begin{minipage}[t]{0.495\textwidth}
\vspace{0pt}
\begin{tcolorbox}[colback=gray!5!white, colframe=black, boxrule=0.5pt, arc=3pt,title=Prompt]
\small Solve X + Y = 5 , 3X - Y = 8. Please reason step by step, and put your final answer within \texttt{\textbackslash{boxed}\{\}}
\end{tcolorbox}
\vspace{-3pt}
\begin{tcolorbox}[colback=gray!10, title={TPV $\alpha=100$ (202 Thinking Tokens)}]
<think> \small To solve the system of equations:
1. Add the two equations to eliminate Y:
  $$
   (X + Y) + (3X - Y) = 5 + 8
   $$
  $$
   4X = 13
   $$
   $$
   X = \frac{13}{4}
  $$

2. Substitute X back into the first equation to find Y:
   $$
   \frac{13}{4} + Y = 5
   $$
   $$
   Y = 5 - \frac{13}{4} = \frac{20}{4} - \frac{13}{4} = \frac{7}{4}
   $$

3. Final Answer:

   $$
   X = \frac{13}{4}, \quad Y = \frac{7}{4}
   $$

</think>

\end{tcolorbox}
\end{minipage}
\hfill
\begin{minipage}[t]{0.4985\textwidth}
\vspace{0pt}
\begin{tcolorbox}[colback=gray!10, title=Original (892 Thinking Tokens)]
<think>
\small Okay, so I have these two equations here: X + Y = 5 and 3X - Y = 8. I need to solve for X and Y. 

Hmm, let me think about how to approach this. 
I remember from class that there are a couple of methods to solve systems of equations, like substitution and elimination. Maybe I can use one of those here.

Let me write down the equations again \ldots

Hmm, \ldots Let me try that.

Adding equation 1 and equation 2:

(X + Y) + (3X - Y) = 5 + 8

Simplify the left side:

X + Y + 3X - Y

Combine like terms:

X + 3X is 4X, and Y - Y is 0. So, 4X = 13

Wait, 5 + 8 is 13, so 4X = 13. Then, to solve for X, I can divide both sides by 4.

X = 13/4

\textbf{Hmm}, \ldots

Okay, now that I have X, I can plug it back into one of the original equations to find Y. 

Let me use equation 1 because it looks simpler: X + Y = 5.

Substituting X = 13/4 into equation 1:

13/4 + Y = 5

To solve for Y, I can subtract 13/4 from both sides.

Y = 5 - 13/4

\textbf{Hmm}, 5 is the same as 20/4, right? So, \ldots

So, Y = 7/4, which is 1.75 or 1 and 3/4.

\textbf{Let me double-check} my work \ldots

Starting with the two equations: \ldots

Which simplifies to \ldots

Then, plugging X back into equation 1: \ldots

Subtract 13/4: \ldots

That also seems correct.

\textbf{Just to be thorough}, let me plug \ldots

\ldots

Yes, that works out. So, both equations are satisfied with X = 13/4 and Y = 7/4.

I think that's it. I used the elimination method by adding the two equations to eliminate Y, solved for X, then substituted back to find Y. Checked my solution in both equations, and it works. 

So, I feel confident that X is 13/4 and Y is 7/4. </think>

\end{tcolorbox}
\end{minipage}%
\end{tcolorbox}
\caption{An example of overthinking through excessive self-verification and redundant, duplicated computation in the thinking process of the base model (right), which is mitigated in the text produced by our model
(bottom left).% result in indecisive Loops (right).  which mitigated show the reasoning trajectory obtained from our models,  which intervention factor of $100$
}
\label{ex:exEquations}
\vspace{-21pt}
}
\end{example}

\end{document}